%% file: main.tex
\pdfoutput=1
\PassOptionsToPackage{numbers, compress}{natbib}
\documentclass[sigconf]{acmart}
\AtBeginDocument{%
  \providecommand\BibTeX{{%
    \normalfont B\kern-0.5em{\scshape i\kern-0.25em b}\kern-0.8em\TeX}}}
    
\usepackage[T1]{fontenc}
\usepackage{aecompl}

\usepackage{tabularx}
\usepackage{makecell}
\usepackage{booktabs}       % professional-quality tables
\usepackage{multirow}
\usepackage{float}
\usepackage{xspace}
\usepackage{hyperref}
\usepackage{algorithm, algpseudocode}

\newcommand{\method}{ResCAL\xspace}

\newcommand{\ie}{{\it i.e.}}
\newcommand{\eg}{{\it e.g.}}

\usepackage{pifont}% http://ctan.org/pkg/pifont

\usepackage{algorithm} % For pseudo-code
\DeclareMathOperator{\argmax}{argmax}
\copyrightyear{2022}
\acmYear{2022}
\setcopyright{acmcopyright}\acmConference[CIKM '22]{Proceedings of the 31st ACM International Conference on Information and Knowledge Management}{October 17--21, 2022}{Atlanta, GA, USA}
\acmBooktitle{Proceedings of the 31st ACM International Conference on Information and Knowledge Management (CIKM '22), October 17--21, 2022, Atlanta, GA, USA}
\acmPrice{15.00}
\acmDOI{10.1145/3511808.3557432}
\acmISBN{978-1-4503-9236-5/22/10}

\begin{document}

%%
%% The "title" command has an optional parameter,
%% allowing the author to define a "short title" to be used in page headers.
\title{Residual Correction in Real-Time Traffic Forecasting}

%%
%% The "author" command and its associated commands are used to define
%% the authors and their affiliations.
%% Of note is the shared affiliation of the first two authors, and the
%% "authornote" and "authornotemark" commands
%% used to denote shared contribution to the research.
\author{Daejin Kim}
\authornote{Both authors contributed equally to this research.}
% \authornote{Equally contributed to this research.}
\email{kiddj@kaist.ac.kr}
\affiliation{%
  \institution{KAIST AI}
  \country{Republic of Korea}
}
% \orcid{1234-5678-9012}

\author{Youngin Cho}
\authornotemark[1]
\email{choyi0521@kaist.ac.kr}
\affiliation{%
  \institution{KAIST AI}
  \country{Republic of Korea}
}

\author{Dongmin Kim}
\email{tommy.dm.kim@kaist.ac.kr}
\affiliation{%
  \institution{KAIST AI}
  \country{Republic of Korea}
%   \streetaddress{P.O. Box 1212}
%   \city{Dublin}
%   \state{Ohio}
%   \country{USA}
%   \postcode{43017-6221}
}

\author{Cheonbok Park}
\email{cbok.park@navercorp.com}
\affiliation{%
  \institution{NAVER Corp.}
  \country{Republic of Korea}
%   \streetaddress{1 Th{\o}rv{\"a}ld Circle}
%   \city{Hekla}
%   \country{Iceland}
}

\author{Jaegul Choo}
\email{jchoo@kaist.ac.kr}
\affiliation{%
  \institution{KAIST AI}
  \country{Republic of Korea}
%   \city{Rocquencourt}
%   \country{France}
}

%%
%% By default, the full list of authors will be used in the page
%% headers. Often, this list is too long, and will overlap
%% other information printed in the page headers. This command allows
%% the author to define a more concise list
%% of authors' names for this purpose.
% \renewcommand{\shortauthors}{Kim, et al.}
\renewcommand{\shortauthors}{Daejin Kim et al.}

%%
%% The abstract is a short summary of the work to be presented in the
%% article.
\input{00_Abstract}

%%
%% The code below is generated by the tool at http://dl.acm.org/ccs.cfm.
%% Please copy and paste the code instead of the example below.
%%


%%
%% Keywords. The author(s) should pick words that accurately describe
%% the work being presented. Separate the keywords with commas.

%% A "teaser" image appears between the author and affiliation
%% information and the body of the document, and typically spans the
%% page.
% \begin{teaserfigure}
%   \includegraphics[width=\textwidth]{sampleteaser}
%   \caption{Seattle Mariners at Spring Training, 2010.}
%   \Description{Enjoying the baseball game from the third-base
%   seats. Ichiro Suzuki preparing to bat.}
%   \label{fig:teaser}
% \end{teaserfigure}

\maketitle

\input{01_Introduction}
\input{02_Related_Works}
\input{03_Preliminaries}
\input{04_Method}
\input{05_Experimental_Settings}
\input{06_Experimental_Results}
\input{07_Conclusion}

%%
%% The acknowledgments section is defined using the "acks" environment
%% (and NOT an unnumbered section). This ensures the proper
%% identification of the section in the article metadata, and the
%% consistent spelling of the heading.
\begin{acks}
% AI 대학원, 중견 (후속), 네이버
This work was supported by the Institute of Information \& communications Technology Planning \& Evaluation (IITP) grant funded by the Korean government (MSIT) (No. 2019-0-00075, Artificial Intelligence Graduate School Program (KAIST)), and the National Research Foundation of Korea (NRF) grant funded by the Korean government (MSIT) (No. NRF-2022R1A2B5B02001913). This work was also partially supported by NAVER Corp.
\end{acks}

%%
%% The next two lines define the bibliography style to be used, and
%% the bibliography file.
\bibliographystyle{./ACM-Reference-Format}
\balance
\bibliography{./reference.bib}

\end{document}

%% file: 00_Abstract.tex
\begin{abstract}

Predicting traffic conditions is tremendously challenging since every road is highly dependent on each other, both spatially and temporally. Recently, to capture this spatial and temporal dependency, specially designed architectures such as graph convolutional networks and temporal convolutional networks have been introduced. While there has been remarkable progress in traffic forecasting, we found that deep-learning-based traffic forecasting models still fail in certain patterns, mainly in event situations (\eg, rapid speed drops). Although it is commonly accepted that these failures are due to unpredictable noise, we found that these failures can be corrected by considering previous failures. Specifically, we observe autocorrelated errors in these failures, which indicates that some predictable information remains. In this study, to capture the correlation of errors, we introduce ResCAL, a residual estimation module for traffic forecasting, as a widely applicable add-on module to existing traffic forecasting models. Our ResCAL calibrates the prediction of the existing models in real time by estimating future errors using previous errors and graph signals. Extensive experiments on METR-LA and PEMS-BAY demonstrate that our ResCAL can correctly capture the correlation of errors and correct the failures of various traffic forecasting models in event situations.

\end{abstract}

%% file: 01_Introduction.tex
\section{Introduction}

\input{Figures/Figure_Concept}

Despite its high practicality, traffic forecasting is a complex task since the speeds of all nodes are highly dominated by their historical signals as well as the conditions of the neighboring nodes.
To handle spatial-temporal datasets, recent studies~\cite{LiYS018, YuYZ18, WuPLJZ19, ZhengFW020, ParkLBTJKKC20, itr2.12044} introduce deep-learning-based models in traffic forecasting and consider the graph structure in the training process.
While these studies have made great progress in the traffic forecasting task, little attention has been given to analyzing the errors in the traffic forecasting models.

In this work, we analyze the errors in the traffic forecasting models and observe that recent models still produce relatively large errors in certain patterns regardless of their high average performance.
While these failures are considered unpredictable, we found that we can estimate \emph{how models will fail} using current errors.
In the real world, it is presumable that correlations among successive errors (\ie, failures) exist. However, previous errors have been ignored in the existing deep-learning-based traffic forecasting methods.
Based on our findings, we highlight the necessity of utilizing previous errors in traffic forecasting.
To explicitly handle the correlation of errors, we utilize historical errors of predictions, \ie, residuals, to make the next prediction. This can correct the failures of the traffic forecasting models and improve their performance in unexpected situations. Consequently, the critical mistakes that are crucial in a real-time setting can be minimized.

In addition to the previous predictions within a single sensor node, the previous predictions of the neighboring nodes are also highly correlated with the current prediction of each sensor node.
To consider both spatial and temporal residual correlations, we propose a simple residual estimation module called \method that estimates the expected residuals in the current prediction, \ie, how forecasting models will fail. 
Our method adopts the spatial-temporal layers conducted with a gated temporal convolutional network (Gated TCN) and a graph convolutional layer (GCN), as proposed in \cite{WuPLJZ19}. Furthermore, we analyze the patterns of failures since high errors occur in certain patterns. To this end, we introduce vector quantization in \method and provide the justification of the calibrations. Vector quantization also allows our method to handle the noise and the outliers that appear in residuals. 

Fig.~\ref{fig_concept} depicts how \method corrects the failures in the real-time setting. Several attempts have been made to consider the residuals in graph classification~\cite{JiaB20, HuangHSLB21}; however, to the best of our knowledge, this study is the first attempt to capture the temporal correlation of residuals in traffic forecasting.
In our experiments, we confirm that the residuals of each node are highly correlated with its previous residuals as well as that of neighboring nodes. Here, we introduce a simple synthetic dataset and validate the correctness of our \method both qualitatively and quantitatively. Subsequently, we conducted extensive experiments in the two most representative traffic datasets: METR-LA and PEMS-BAY. In both datasets, we calibrated the predictions from various traffic forecasting models such as STGCN~\cite{YuYZ18}, DCRNN~\cite{LiYS018}, Graph WaveNet~\cite{WuPLJZ19}, and STAWnet~\cite{itr2.12044}. 
Specifically, we focus on whether our \method accurately calibrates the predictions in the time areas where the existing models struggle. Despite its simplicity, our \method consistently improves performance in event situations. 
Along with correcting failures, we validate the effectiveness of the vector quantization approach with a qualitative analysis. 
In this analysis, we verify that our \method can provide meaningful justification for the calibration by examining the residual patterns in the unobserved data.

Our contributions can be summarized as follows:
\begin{itemize}
    \item We highlight the importance of utilizing the historical predictions to explicitly handle the correlation among errors that occur in a real-time setting. 
    \item We propose a novel method called \method as a widely-applicable add-on module, which estimates the future residuals and corrects the failures of models.
    % \item We conducted extensive experiments to demonstrate that our \method consistently improves the baselines in event situations while significantly reducing the correlation of residuals.
    \item Extensive experiments demonstrate that our \method consistently improves the baselines in event situations while significantly reducing the correlation of residuals.
\end{itemize}

%% file: Figures/Figure_Concept.tex
\begin{figure}[t!]
\begin{center}
    \includegraphics[width=1.0\linewidth]{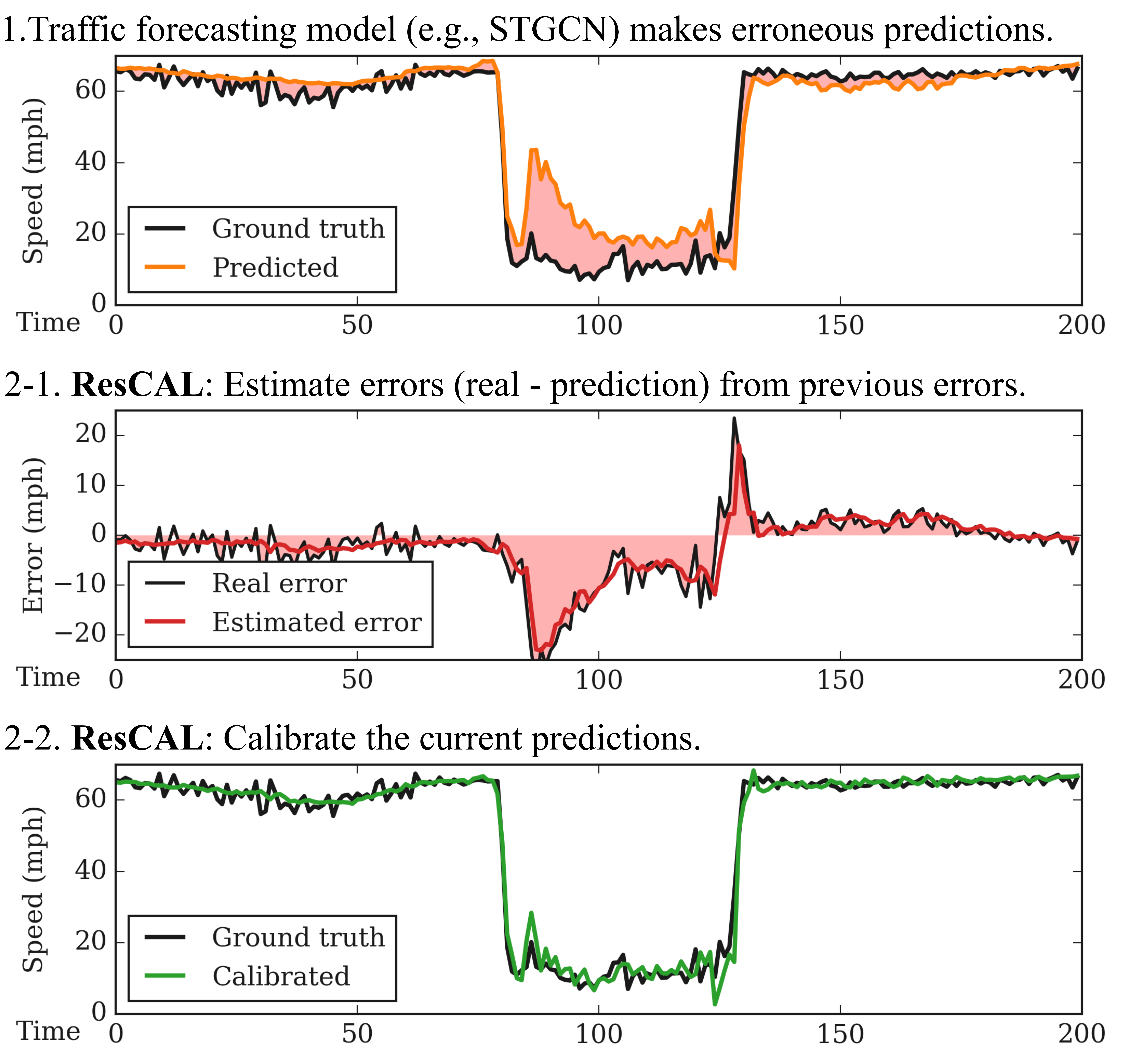}
\end{center}
    \vspace{-0.1in}
    \caption{Using the pretrained traffic forecasting model, our proposed \method calibrates the predictions by estimating the errors (residuals), \ie, failures of a model, and further improves the prediction performance. In traffic forecasting, predicting future errors is quite feasible since the previous errors from the model can be correlated with the current prediction.}
\label{fig_concept}
\vspace{-0.1in}
\end{figure}

%% file: 02_Related_Works.tex
\section{Related Works}
\label{sec_related_works}

\input{Figures/Figure_Correlation}

\subsection{Traffic Forecasting}
Traffic forecasting is a challenging task due to the complicated spatial and temporal dependencies among the sensor nodes. To capture the dynamics of traffic conditions, data-driven approaches based on deep learning have received considerable attention. In spatial and temporal modeling, approaches based on graph convolutional networks (GCNs)~\cite{KipfW17,ZhangSXMKY18,AtwoodT16} are promising in capturing spatial dependencies among roads. 
Several studies~\cite{ZhangZQLY16, ChengZZX18, WuT16} have proposed applying recurrent neural networks (RNNs) and convolutional neural networks (CNNs) to capture the temporal dependencies along the sequence in traffic forecasting.

Recent studies~\cite{LiYS018, YuYZ18, WuPLJZ19, ZhengFW020, ParkLBTJKKC20, itr2.12044} have shown that graph modeling is a key factor to achieve state-of-the-art performance in traffic forecasting.
Particularly, DCRNN~\cite{LiYS018} demonstrates its impressive performance against statistical approaches by incorporating diffusion graph convolutional neural networks~\cite{AtwoodT16} into RNNs.
STGCN~\cite{YuYZ18} utilizes only convolution-based approaches for modeling spatial and temporal dependencies in the road network. Graph WaveNet~\cite{WuPLJZ19} introduces a self-adaptive adjacency matrix to overcome the limitation of applying fixed adjacency information and uses dilated convolutions~\cite{OordDZSVGKSK16} to efficiently handle long sequences. 
As another approach, STAWnet~\cite{itr2.12044} applies a self-attention mechanism~\cite{VaswaniSPUJGKP17} to capture spatial dependencies between roads and uses self-learned node embedding to eliminate the need of prior knowledge of the graph structure.
While these studies have shown remarkable progress in solving the traffic forecasting problem, methods to consider the errors of the forecasting models have been under-explored. In the line of traffic forecasting research, our work can improve the performance of the existing forecasting models by calibrating predictions of those models in real time.

\subsection{Residual Correction}
In statistical approaches, residual is defined as the difference between the ground truth values and the predicted ones.
Statistical time-series models such as the autoregressive model and moving-average model represent future values as a linear combination of observed values and residuals at previous times steps. The autoregressive integrated moving average (ARIMA)~\cite{box1976time} model considers residuals with autoregressive terms. 
However, these approaches suffer from expressing non-linearity because the residual term is expressed as a finite linear combination of a white-noise sequence.

Fitting residuals in the regression problem is an effective technique in machine learning.
For example, a gradient boosting algorithm~\cite{friedman2001greedy} and its variants, such as XGBoost~\cite{chen2015xgboost} and LightGBM~\cite{KeMFWCMYL17}, recursively capture the residuals to improve the performance.
% to improve the prediction accuracy.
Moreover, graph neural network architectures to model residual correlation have also been proposed.
% in previous research.
\cite{JiaB20} suggested modeling of the joint distribution of the residuals to obtain information from both the input feature and the output correlation in the graph structure. \cite{HuangHSLB21} proposes a procedure called “Correct and Smooth”, which models error correlation to correct the base prediction from simple architectures such as multi-layer perceptron (MLP). In contrast to these works, our \method considers both spatial and temporal residuals to enhance the performance of traffic forecasting models.

\subsection{Discrete Representations}

Utilizing the discrete representation can provide interpretability and reduce the noise for a given data~\cite{ChenCDHSSA16, OordVK17, JangGP17}. 
One of the discretization approaches is vector quantization~\cite{gray1984vector}, which is suggested as a system for mapping a signal into a digital sequence. 
VQ-VAE~\cite{OordVK17} utilizes this quantization mechanism to model the categorical distribution, and the latent variable is represented as the combination of the embedding vectors. 
The vector quantization layer formulates a latent space, called codebook, and clusters vectors according to a given distance metric (\eg, L2 distance). 
To learn a discrete representation by the backpropagation algorithm, \cite{JangGP17} suggests Gumbel-Softmax which can approximate categorical samples by a differentiable sampling mechanism. 
In this work, we utilize discrete representation to enhance the interpretability of the calibration module and to reduce the noise introduced in traffic forecasting.

%% file: Figures/Figure_Correlation.tex
\begin{figure*}[t!]
\begin{center}
    \includegraphics[width=1.0\linewidth]{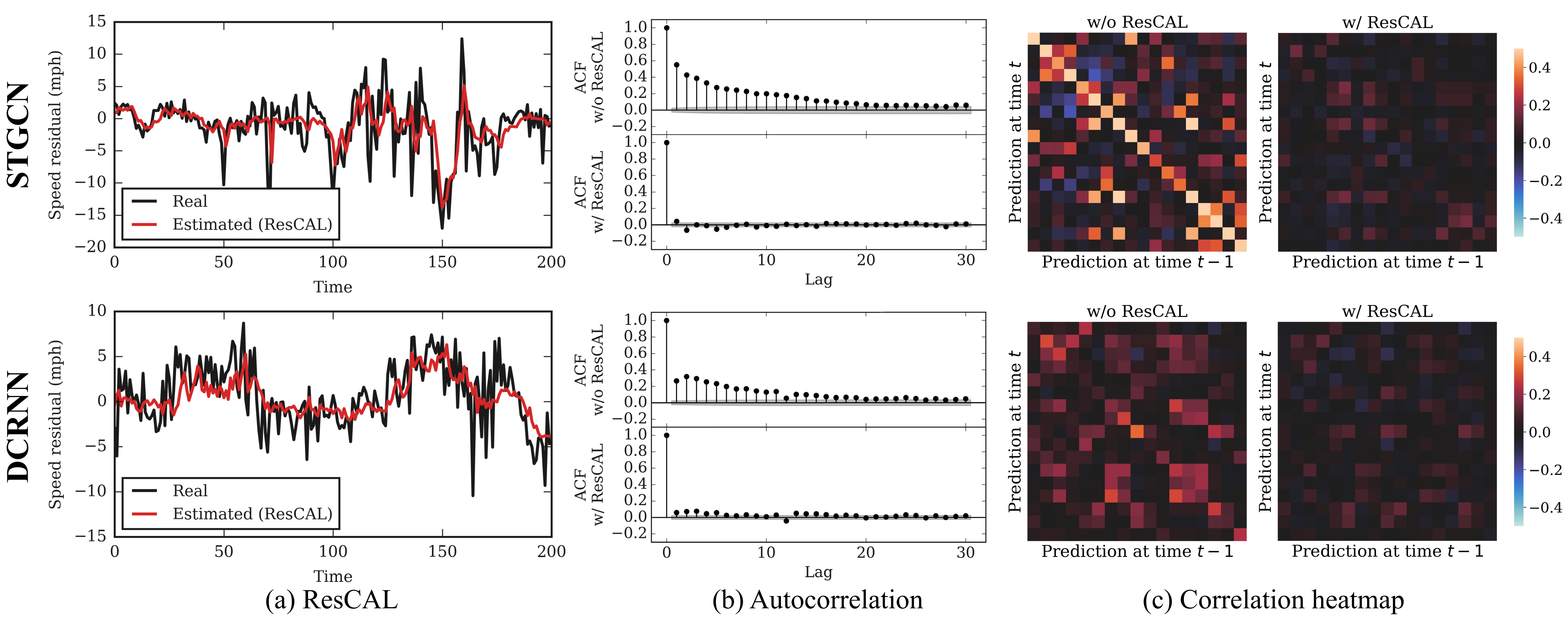}
\end{center}
    \vspace{-0.1in}
    \caption{Reduction in the autocorrelation of residuals on the METR-LA dataset. We analyzed the 5 minutes ahead prediction results on node 19, and node 22 for STGCN, and DCRNN, respectively.
    (a) plots the ground truth residual and the estimated residual for each case, and shows that our proposed \method accurately estimates the residual for both cases.
    (b) represents the ACF plots of residuals with and without the calibration of \method. After applying \method, the autocorrelation at every lag decreases to almost zero, indicating temporal dependencies are captured by \method.
    (c) shows the heatmaps of lag 1 autocorrelation of residuals on a neighborhood of node 19 for STGCN, and node 22 for DCRNN. $(i, j)$-th element of the heatmap shows a Pearson Correlation of residuals between the $i$-th node at time $t$ and $j$-th node at time $t-1$. The brighter point represents a high correlation between the two corresponding nodes. For both cases, \method drastically reduces the correlation between these nodes, indicating that \method can capture both temporal and spatial dependencies to correct the predictions.
    }
\label{fig_correlation_all}
\vspace{-0.1in}
\end{figure*}

%% file: 03_Preliminaries.tex
\section{Residual Diagnostics in Traffic Forecasting}

\noindent
\textbf{High Errors in Traffic Forecasting.} In the real world, a few critical errors have a huge influence on traffic conditions. Therefore, predicting such errors is essential for the traffic forecasting task.
On a widely used traffic dataset, METR-LA~\cite{LiYS018}, we observed that top-$20\%$ errors account for about $60\%$ of the total absolute errors, \eg, $63.0\%$ for STGCN and $62.8\%$ for DCRNN.
In traffic dynamics, these abnormal cases are mainly caused by sparse events.
In this work, we specifically focus on time steps with top-$20\%$ errors and denote them as \emph{event situations} since a large magnitude of error indicates the failure of the prediction.

\noindent
\textbf{Residual Autocorrelation in Traffic Forecasting.} 
Autocorrelation is the correlation of a time series and its delayed copy. Given an input sequence $y_1, y_2, ..., y_T$, the autocorrelation function (ACF) $r_k$ measures the degree of the linear relationship between $y_{t}$ and $y_{t-k}$ where $k$ is a time lag:
\begin{equation}
    r_{k} = \frac{\sum_{t=k+1}^{T}(y_{t}-\bar{y})(y_{t-k}-\bar{y})}{\sum_{t=1}^{T}(y_{t} - \bar{y})^2}.
\end{equation}
High autocorrelation indicates a high potential of performance improvement; any forecasting model of which residuals are correlated or residuals have a non-zero mean can be improved by estimating the future residuals~\cite{hyndman2018forecasting}.
The residuals of the traffic forecasting model are commonly autocorrelated; therefore, we explicitly capture this relationship to further enhance the model performance.

Fig.~\ref{fig_correlation_all}~(b) shows the ACF of residuals for the traffic forecasting models. Before calibration, the residuals have a high autocorrelation, meaning that predictable information remains in the residuals.
Fig.~\ref{fig_correlation_all}~(c) represents the Pearson Correlation between the current residual of each sensor node and the previous residuals of neighboring sensor nodes on METR-LA. The bright points in the heatmap of the existing forecasting models show that the residuals in the neighboring sensor nodes are highly correlated. 
By calibrating the predictions, the correlation among the residuals can be significantly reduced, as shown in the ACF plot and the heatmap. 

%% file: 04_Method.tex
\section{Residual Correction in Traffic Forecasting}
\label{sec_method}

\input{Figures/Figure_Residuals}
\input{Figures/Figure_Method}

In this section, we introduce a simple model-agnostic framework to boost the performance of traffic forecasting models in the real-time setting.
To this end, we first formally describe the problem setting considered in traffic forecasting. Next, the model architecture and the residual prediction mechanism of our \method are introduced.

\subsection{Real-Time Traffic Forecasting} \label{problem_definition}

The aim of traffic forecasting is to predict the future traffic speed observed at $N$ correlated sensors on the road network. Let $\mathcal{G}=(\mathcal{V},\mathcal{E})$ be a graph representing the spatial relationship between sensors where $V$ is a set of nodes ($|\mathcal{V}|=N$) and $\mathcal{E}$ is a set of edges. Following the convention of the traffic prediction problem, $X^{t}\in \mathbb{R}^{N\times 2}$ is denoted as the graph signal obtained at time $t$; $X^{t}_{:,1}$ and $X^{t}_{:,2}$ represent the speed and the timestamp features, respectively. The goal of traffic forecasting is to learn a mapping function $f$ from past graph signals and a graph $\mathcal{G}$ to future traffic speeds:
\begin{equation}
    [X^{(t-T_x+1):t}, \mathcal{G}] \xrightarrow[]{f} X^{(t+1):(t+T_y)}_{:,1},
\end{equation}
where $T_x$ and $T_y$ are the input and output sequence lengths, respectively. 
In particular, we consider a real-time traffic forecasting problem in which the current prediction of the model can be corrected using continuous historical predictions. Let $Y^t=X^{(t+1):(t+T_y)}_{:,1} \in \mathbb{R}^{N\times T_y}$ be the ground truth speed so that $Y^t_{:,i}$ represents the ground truth speed of $i$ step ahead prediction at time $t$, and $\hat{Y}^t$ is its estimated values, \ie, $\hat{Y}^t=f(X^{(t-T_x+1):t}, \mathcal{G})$. Then, the residual of traffic prediction is defined as $R^t=Y^t-\hat{Y}^t$. Note that the ground truth of $R^t_{:,i}$ can be observed when time is greater than or equal to $t+i$. $U^t=[R^{(t-1)}_{:,1}, R^{(t-2)}_{:,2}, ..., R^{(t-T_y)}_{:,T_y}]\in \mathbb{R}^{N \times T_y}$ is denoted as the newly observed residuals at time $t$. Here, our main problem is learning a mapping function $g$ which predicts the residual at time $t$, given past graph signals, observed residuals, and a graph $\mathcal{G}$:
\begin{equation}
    [X^{(t-T_x+1):t}, U^{(t-T_u+1):t}, \mathcal{G}] \xrightarrow[]{g} R^t,
\end{equation}
where $T_u$ is the time steps of the observed residuals. Given the estimated residuals $\hat{R}^t=g(X^{(t-T_x+1):t}, U^{(t-T_u+1):t}, G)$, accurate predictions can be made by taking $\hat{Y}^t+\hat{R}^t$ as a final output. Fig.~\ref{fig_residuals} illustrates our problem setting.

% Considering the real-time traffic forecasting problem has several advantages:
Our proposed problem setting in real-time traffic forecasting has several advantages, as follows:
(i) The correlation between successive residuals can be explicitly modeled. We may consider resizing the window size of input to handle this correlation in the time series model. However, this can be tricky depending on the model design and is time-consuming to retrain due to its high complexity.
(ii) As most parts of the complex modeling is done by $f$, we can learn $g$ with a relatively lightweight model to estimate the residuals. This allows hyperparameter tuning with a small budget and increases the reusability of the model in real-world situations.
(iii) Since information about $f$ is not needed to predict the residuals, the performance of the base forecasting model can be improved in a model-agnostic way.

\subsection{\method}

To explicitly model the residuals, we designed a light-version of the traffic forecasting model called \method consisting of spatial-temporal layers with the self-adjacency matrix, as proposed by \cite{WuPLJZ19}.
The overall architecture of our \method is depicted in Fig.~\ref{fig_method}. The encoder consists of $k$ spatial-temporal layers, each conducted by a gated temporal convolutional layer (Gated TCN) and a graph convolutional layer (GCN).
Following the conventional setting of the base traffic forecasting models, we let $T=T_x=T_y$. The encoder produces a latent $Z\in\mathbb{R}^{N \times d_h}$ by taking both a graph signal $X\in\mathbb{R}^{N \times 2 \times T}$ and a residual $U\in\mathbb{R}^{N \times T_u \times T}$:
\begin{equation}
    Z=\textbf{Encoder}(\textbf{Concat}(X, U)),
\end{equation}
where $\textbf{Concat}$ is a concatenation operation along the second axis and $d_h$ is the number of hidden dimensions.

Here, we want to make a more accurate prediction by capturing useful information from $Z$ and simultaneously provide a reason for the judgment. However, it is difficult to directly analyze $Z$ since we do not put any strict restrictions on generating $Z$. Moreover, the noise introduced by the baseline model or data further hinders the analysis of $Z$. 
Previous works~\cite{OordVK17, JangGP17} introduce discrete latent vectors into an autoencoder to increase the interpretability and stabilize training on the noisy data. However, we observe that using only these discrete forms sometimes reduces overall performance of the model. Instead, we use a hybrid approach of combining discrete and continuous representations. 
We take the sum of the outputs of two different branches from $Z$: the regression branch and the quantization branch.
The regression branch provides a continuous representation of the accurate estimation of the residual while the quantization branch provides interpretable and denoised information using discrete representation. The Straight-Through (ST) Gumbel estimator~\cite{JangGP17} is used to provide the differentiable discrete variable $y \in \{0,1\}^k$ given an input $x\in \mathbb{R}^k$: 
\begin{equation}
\begin{aligned}
y &= \textbf{ST-Gumbel}(x)\\
&= \textbf{one\_hot}(\argmax_i [g_i + x_i]),
\end{aligned}
\end{equation}
where $g_1, g_2, ..., g_k$ are i.i.d samples drawn from the standard Gumbel distribution. This variable is smoothly approximated in the backward pass.
Formally, with the learnable embedding vector $E\in \mathbb{R}^{d_e \times {(d_c \times n_c)}}$, the quantized vector $Q\in \mathbb{R}^{N \times {(d_c \times n_c)}}$, the estimated future residual $R\in \mathbb{R}^{N \times T_y}$, and the output of \method is calculated as:
\begin{equation}
\begin{aligned}
W &= f_q(Z), \\
R &= f_o(f_r(Z) + QE^{\top}), \\
Q_{i,j_k+1:j_k+n_c} &= \textbf{ST-Gumbel}(W_{i,j_k+1:j_k+n_c}),
\end{aligned}
\end{equation}
where $n_c$ is the number of categories, $d_c$ is the number of categorical variables, $d_e$ is the dimension of the embedding matrix, and $j_k:=k\cdot n_c$ for $k=0,...,d_c-1$; $f_o$, $f_r$, and $f_q$ are the output layer, regression layer, and quantization layer, respectively. Each layer consists of a combination of a pointwise convolution and a ReLU activation.

By predicting the residuals, we can easily calibrate the models and further improve the existing traffic forecasting models. In real-time prediction, the previous residuals emerged by the prediction model as well as the current traffic data are used as the input and the future residuals are predicted. The clustering results can also be used to interpret the behavior of the model. In Section~\ref{sec_exp_results}, we will show that the baselines fail similarly for similar events. Using the predicted residuals, we can now calibrate the output of the prediction model. By calibrating the future prediction with the current errors, we can consider the temporal dependency between residuals in traffic forecasting. Our experiments show that such improvements cannot be achieved solely by increasing the capacity of the base forecasting model with additional parameters or utilizing longer sequences of the input without considering the temporal dependencies of the residuals as our \method does.

%% file: Figures/Figure_Residuals.tex
\begin{figure}[t!]
\begin{center}
    \includegraphics[width=0.95\linewidth]{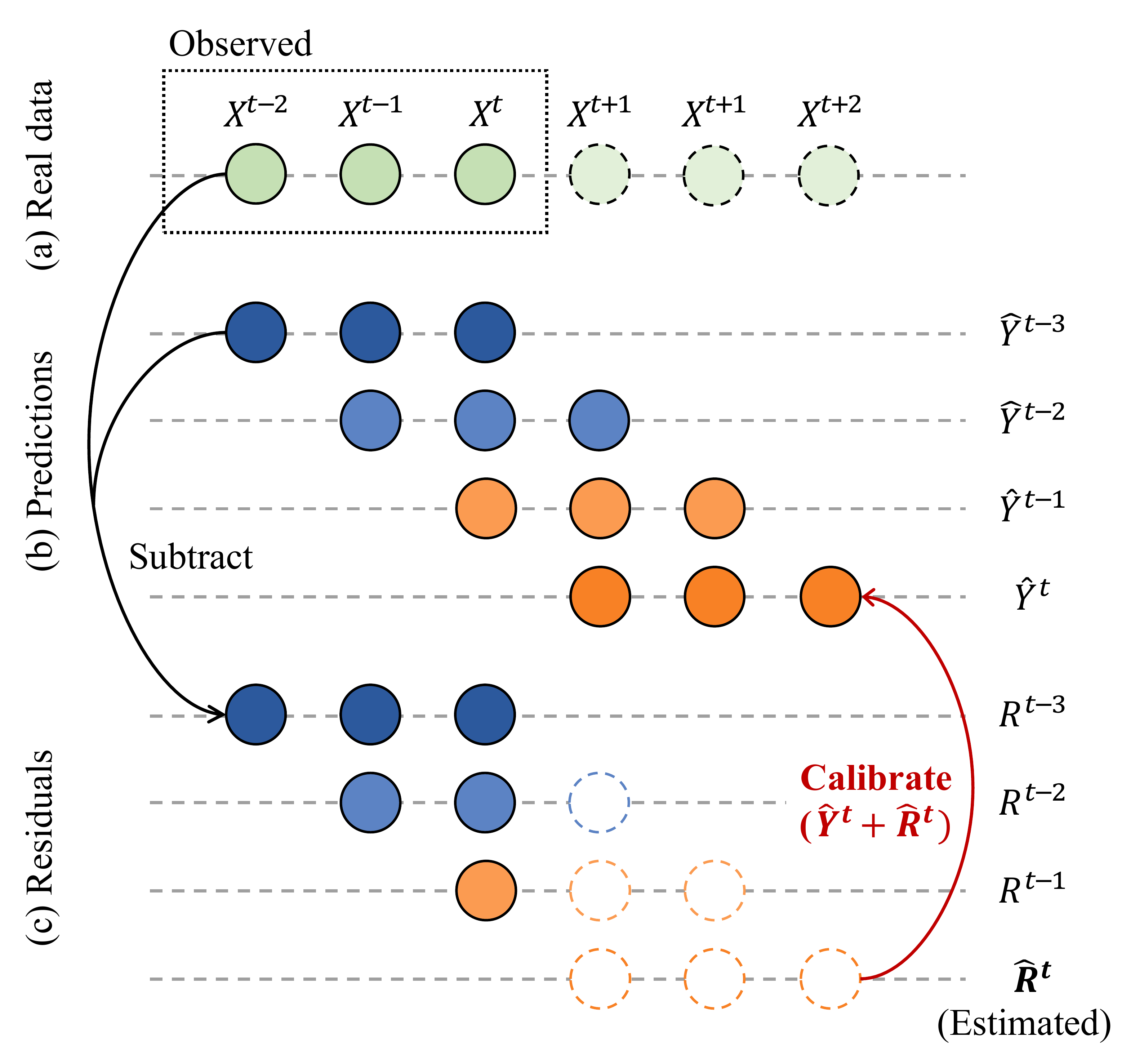}
\end{center}
    \vspace{-0.2in}
    \caption{ Estimation of residuals at time $t$. Residuals (c) available at time $t$ are based on previous predictions (b) and observed real data (a). The aim of \method is to make an estimation on the residuals for the current prediction.}
\label{fig_residuals}
\vspace{-0.1in}
\end{figure}

%% file: Figures/Figure_Method.tex
\begin{figure*}[t!]
\begin{center}
    \includegraphics[width=1.0\linewidth]{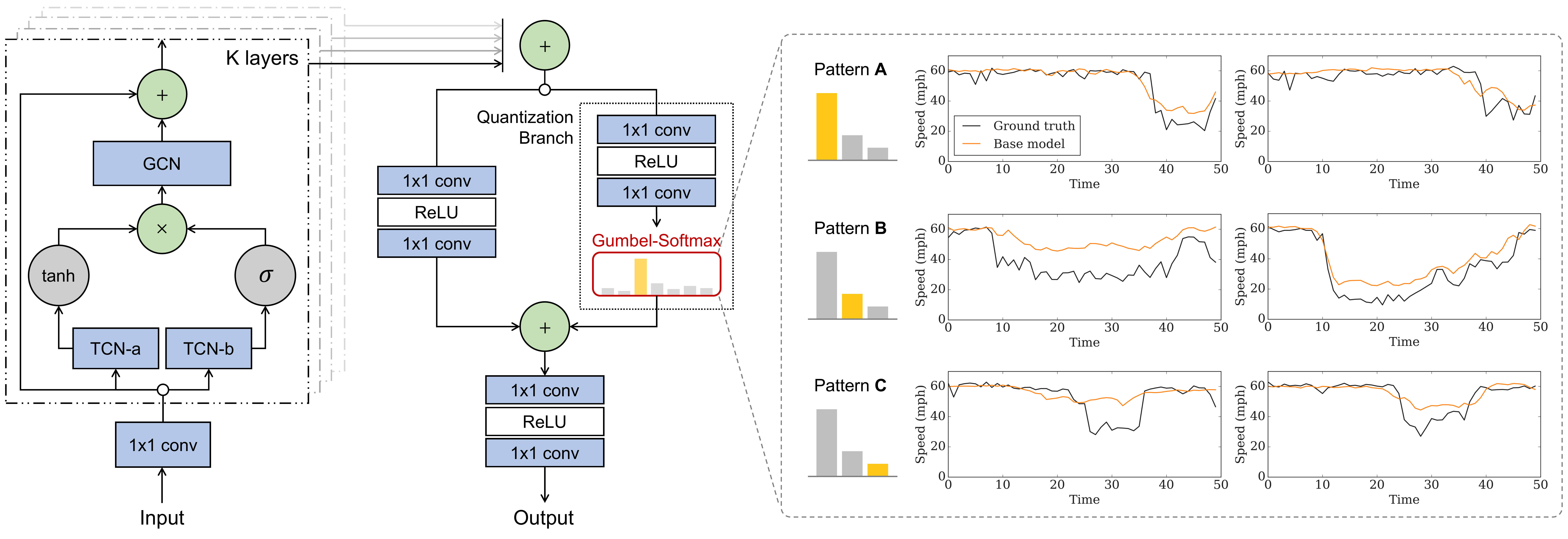}
\end{center}
    \vspace{-0.2in}
    \caption{The overall architecture of our \method. The encoder consists of spatial-temporal layers and the decoder with the quantization branch outputs the residuals expected to occur in the following predictions. The quantization branch with Gumbel-Softmax operation classifies the patterns of events and provides the interpretation of the events and the failures of the base models.}
\label{fig_method}
\vspace{-0.1in}
\end{figure*}

%% file: 05_Experimental_Settings.tex
\section{Experimental Settings}
\label{sec_exp_settings}

\input{Tables/Table_Dataset}
\input{Figures/Figure_Dataset_Map}

\input{Figures/Figure_Qual_Main_Results}

\subsection{Traffic Dataset}

We examined our \method on the two representative traffic datasets: PEMS-BAY and METR-LA introduced by \cite{LiYS018}.
PEMS-BAY, collected by California Transportation Agencies (CalTrans) Performance Measurement System (PeMS), contains 325 sensor data from the Bay Area. In our experiments, 6 months of data collected from Jan 1st, 2017 to May 31st, 2017 was selected.
METR-LA contains the traffic speed data recorded by 207 sensors on the highways of Los Angeles County~\cite{JagadishGLPPRS14}.
The sensor locations of METR-LA and PEMS-BAY are shown in Fig.~\ref{fig_dataset_map}.
Both datasets were preprocessed following \cite{LiYS018}. For METR-LA, 4 months of data collected from Mar 1st, 2012 to Jun 30th, 2012 was selected for our experiments.
For both datasets, the traffic speed is aggregated into 5 minute windows, and Z-score normalization is applied to the input. As \cite{ShumanNFOV13} proposed, adjacency matrix of both datasets was conducted using road distances with a threshold Gaussian kernel.
A detailed description of each dataset is provided in Table~\ref{table_dataset}.

\subsection{Baseline Methods}
\label{baselines}
To widely validate the correctness and effectiveness of our \method, we examined several baselines commonly used for traffic forecasting. 
We additionally provide the reported results for the basic models such as ARIMA and FC-LSTM to help compare the performance of various models.
In this work, we choose the mean squared error (MSE) for training the base models. 
For all baselines, the PyTorch implementation was utilized as denoted in their footnotes, and their default training strategies were followed.

\begin{itemize}
\item \textbf{ARIMA.} The auto-regressive integrated moving average model with a Kalman filter, which is the most representative regression model for time series data.
 
\item \textbf{FC-LSTM~\cite{SutskeverVL14}.} Basic deep-learning-based regression model for time series data conducted with long short-term memory (LSTM)~\cite{HochreiterS97} and fully-connected layers.

\item \textbf{DCRNN~\cite{LiYS018}.}\footnote{\url{github.com/chnsh/DCRNN_PyTorch}} Diffusion convolutional recurrent neural network consisting of the graph convolutional networks and recurrent neural networks.

\item \textbf{STGCN~\cite{YuYZ18}.}\footnote{\url{github.com/FelixOpolka/STGCN-PyTorch}} Spatial-temporal graph convolutional network which is conducted with the graph convolutional layer and 1D convolutional layers.

\item \textbf{Graph WaveNet~\cite{WuPLJZ19}.}\footnote{\url{github.com/nnzhan/Graph-WaveNet}} Traffic forecasting model which combines the dilated 1D convolutional layers and graph convolutional networks.

\item \textbf{STAWnet~\cite{itr2.12044}.}\footnote{\url{github.com/CYBruce/STAWnet}} Spatial-temporal attention network with temporal convolution and spatial attention mechanism to capture dynamic spatial dependencies.

\end{itemize}

\input{Tables/Table_Synthetic}
\input{Tables/Table_High_Error}

\subsection{Training Settings}

For both METR-LA and PEMS-BAY, we used the same training strategy for the simplicity. $k=4$ spatial-temporal layers were used for the encoder, and  $d_c=32$, $n_c=16$, and $d_e=16$ were set for the quantization branch in the decoder.
Since the baselines predict the next $T=12$ steps in units of 5 minutes and each step has errors of 12 horizons, the input residuals and predictions have a size of $12 \times 12$ and $12 \times 1$, respectively.
Our \method is trained with the mean absolute error (MAE) and a batch size of 256. An Adam optimizer with a learning rate of 0.001, $\beta_{1} =0.9$ and $\beta_{2} =0.999$ is also used.
Each dataset is split into a training set, validation set, and test set with a ratio of 7:1:2 and the model with the best validation score is selected in all experimental evaluations.

%% file: Tables/Table_Dataset.tex
\begin{table}[t!]
\begin{center}
\caption{Detailed description of the synthetic data, METR-LA and PEMS-BAY used in our experiments.}
\vspace{-0.1in}
\label{table_dataset}
\small
\begin{tabular}[\linewidth]{lccccc}
\toprule
Data & \# Nodes    & \# Edges  & \# Time steps \\
\midrule
Synthetic   & -    & -  & 10000 \\
METR-LA     &   207 & 1515  & 34272 \\
PEMS-BAY    &   325 & 2369  & 52116 \\
\bottomrule
\end{tabular}
\end{center}
\vspace{-0.1in}
\end{table}

%% file: Figures/Figure_Dataset_Map.tex
\begin{figure}[t!]
\begin{center}
    \vspace{-0.05in}
    \includegraphics[width=0.7\linewidth]{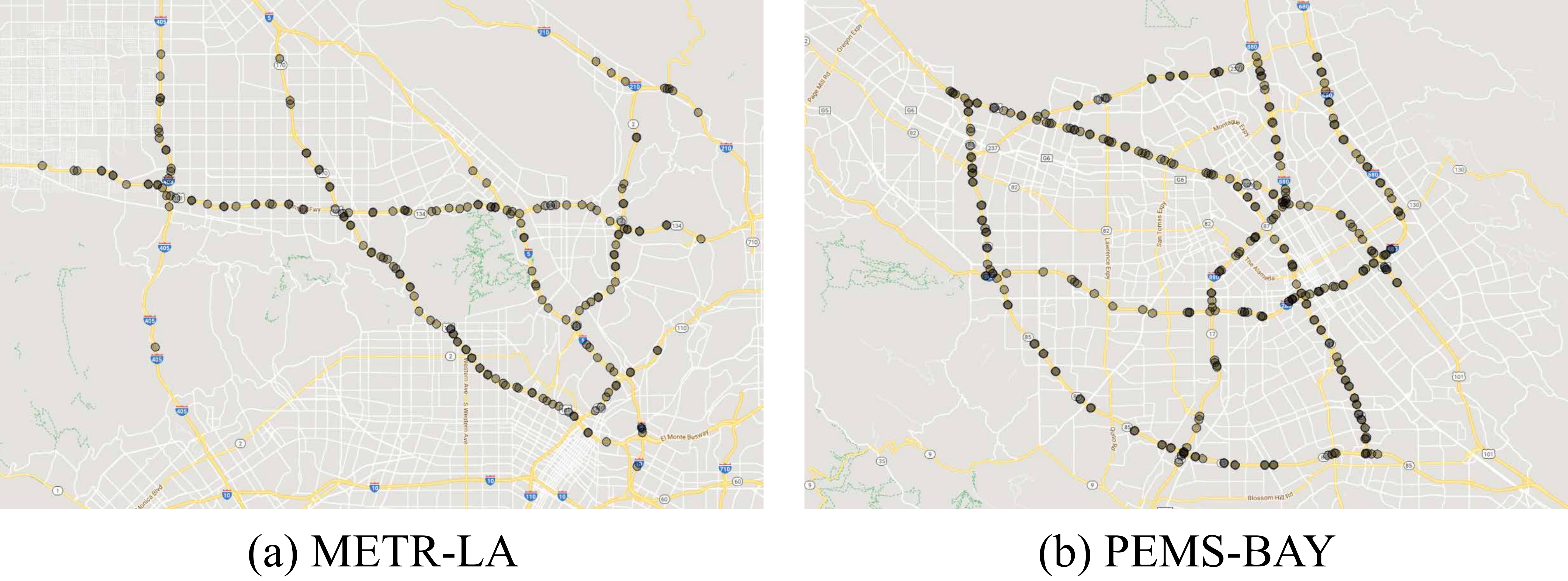}
\end{center}
    \vspace{-0.15in}
    \caption{\mbox{Sensor locations of (a) METR-LA and (b) PEMS-BAY.}}
    \vspace{-0.2in}
\label{fig_dataset_map}
\end{figure}

%% file: Figures/Figure_Qual_Main_Results.tex
\begin{figure*}[t!]
\begin{center}
    \includegraphics[width=1.0\linewidth]{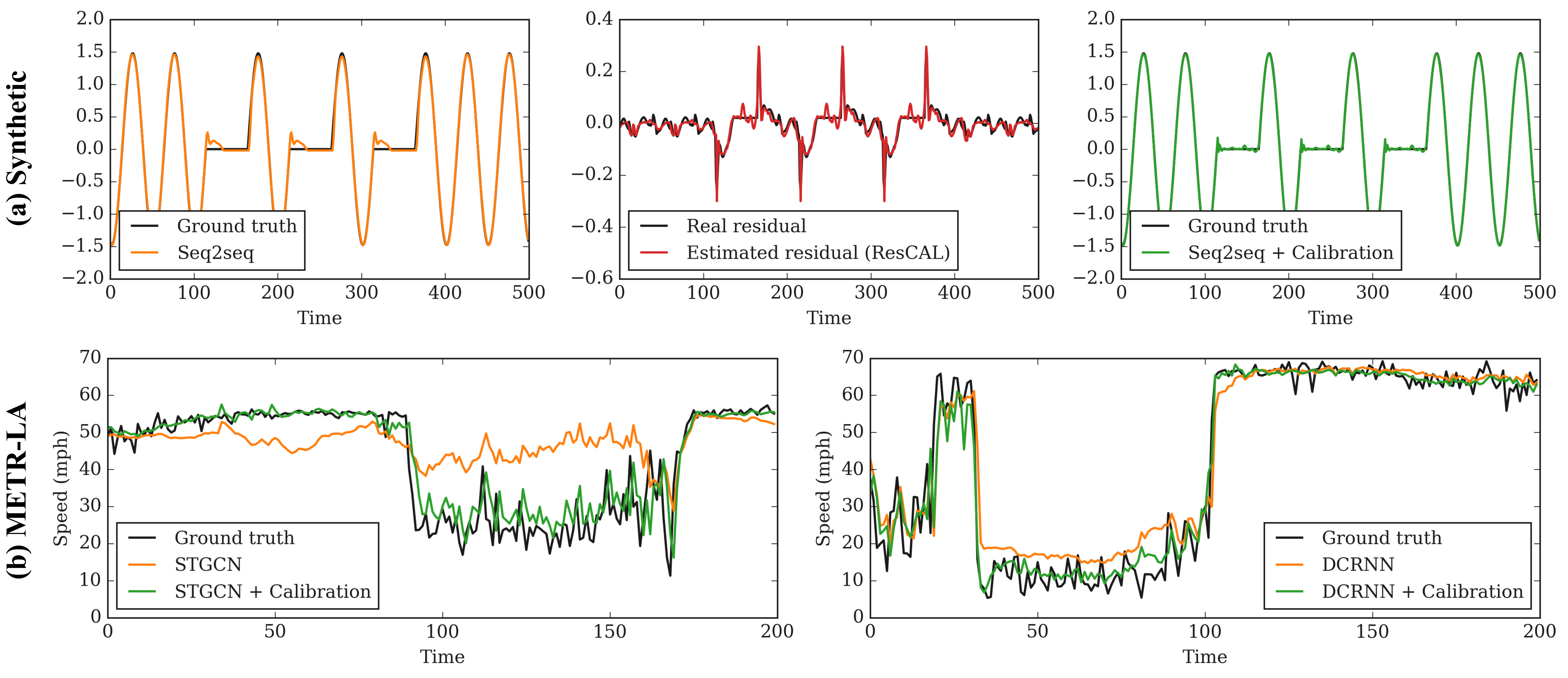}
\end{center}
\vspace{-0.1in}
    \caption{Calibrated predictions using \method. 
    (a) On the synthetic dataset, Seq2Seq shows similar failures for each event, and \method correctly calibrates the predictions. 
    % similarly fails at predicting the event patterns. 
    (b) On METR-LA, \method accurately corrects the failures of the baseline models.}
    % \vspace{-0.1in}
\label{fig_qual_main_results}
\end{figure*}

%% file: Tables/Table_Synthetic.tex
\begin{table}[t!]
\begin{center}
\caption{Experimental results on the synthetic dataset for LSTM Seq2seq with and without our \method.}
\vspace{-0.1in}
\label{table_synthetic}
\small
\begin{tabular}[\linewidth]{llccc}
\toprule
Time steps & Models    & MAE  & RMSE   & MAPE \\
\midrule\midrule
\multirow{2}{*}{1 step}
& Seq2seq & 0.029  & 0.047 & 7.37\% \\
& \textbf{+ Calibration} & \textbf{0.015}  & \textbf{0.022} & \textbf{2.95\%} \\
\midrule
\multirow{2}{*}{6 step}
& Seq2seq & 0.061  & 0.153 & 20.34\% \\
& \textbf{+ Calibration} & \textbf{0.025}  & \textbf{0.093} & \textbf{10.29\%} \\
\midrule
\multirow{2}{*}{12 step}
& Seq2seq & 0.123  & 0.294 & 51.70\% \\
& \textbf{+ Calibration} & \textbf{0.067}  & \textbf{0.205} & \textbf{27.56\%} \\
\midrule
\multirow{2}{*}{24 step}
& Seq2seq & 0.174  & 0.367 & 80.64\%\\
& \textbf{+ Calibration} & \textbf{0.122}  & \textbf{0.291} & \textbf{53.47\%} \\
\bottomrule
\end{tabular}
\vspace{-0.1in}
\end{center}
\end{table}

%% file: Tables/Table_High_Error.tex
\begin{table*}[t!]
\begin{center}
\caption{Quantitative results in event situations where the absolute error of the base forecasting model falls within the top $20\%$.
The performances of the forecasting models were measured with and without our \method on the METR-LA and PEMS-BAY datasets. 
The results are reproduced as mentioned in Section~\ref{sec_exp_settings}.
Note that the high error region depends on the base forecasting model.
}
\label{table_high_error}
\small
\begin{tabular}[\linewidth]{l l ccc | ccc | ccc | ccc}
\toprule
\multirow{2}{*}{Data}    & \multirow{2}{*}{Models}  & \multicolumn{3}{c}{15min}   & \multicolumn{3}{c}{30min} & \multicolumn{3}{c}{60min}   & \multicolumn{3}{c}{Average}  \\
\cmidrule{3-14}
&&  MAE & RMSE  & MAPE  &  MAE & RMSE  & MAPE  &  MAE & RMSE  & MAPE &  MAE & RMSE  & MAPE \\

\midrule\midrule

\multirow{9}{*}{\rotatebox[origin=c]{90}{METR-LA}}
& DCRNN~\cite{LiYS018} & 14.53 & 16.31 & 48.63\% & 16.98 & 18.93 & 60.79\% & 20.03 & 22.00 & 75.27\% & 17.18 & 19.08 & 61.56\% \\ 
& \textbf{+ Calibration} & \textbf{13.46} & \textbf{15.44} & \textbf{44.22\%} & \textbf{16.26} & \textbf{18.38} & \textbf{57.37\%} & \textbf{19.45} & \textbf{21.54} & \textbf{73.13\%} & \textbf{16.39} & \textbf{18.46} & \textbf{58.24\%} \\ 
\cmidrule{3-14}
& STGCN~\cite{YuYZ18} & 15.29 & 16.93 & 53.19\% & 17.94 & 19.78 & 65.91\% & 21.17 & 23.00 & 82.76\% & 18.13 & 19.90 & 67.29\% \\ 
& \textbf{+ Calibration} & \textbf{12.77} & \textbf{15.07} & \textbf{43.55\%} & \textbf{15.79} & \textbf{18.18} & \textbf{57.50\%} & \textbf{19.10} & \textbf{21.44} & \textbf{73.93\%} & \textbf{15.88} & \textbf{18.23} & \textbf{58.33\%} \\ 
\cmidrule{3-14}
& Graph WaveNet~\cite{WuPLJZ19} & 13.39 & 15.06 & 41.49\% & 16.15 & 17.99 & 54.92\% & 19.08 & 20.93 & 69.24\% & 16.21 & 17.99 & 55.22\% \\ 
& \textbf{+ Calibration} & \textbf{13.28} & \textbf{14.99} & \textbf{41.45\%} & \textbf{16.03} & \textbf{17.93} & \textbf{54.43\%} & \textbf{18.94} & \textbf{20.85} & \textbf{68.34\%} & \textbf{16.08} & \textbf{17.93} & \textbf{54.74\%} \\ 
\cmidrule{3-14}
& STAWnet~\cite{itr2.12044} & 13.56 & 15.30 & 43.75\% & 16.15 & 18.10 & 56.00\% & 19.00 & 21.00 & 68.44\% & 16.24 & 18.13 & 56.06\% \\ 
& \textbf{+ Calibration} & \textbf{13.17} & \textbf{14.98} & \textbf{42.24\%} & \textbf{15.87} & \textbf{17.89} & \textbf{54.88\%} & \textbf{18.80} & \textbf{20.84} & \textbf{68.21\%} & \textbf{15.95} & \textbf{17.90} & \textbf{55.11\%} \\ 

\midrule\midrule

\multirow{9}{*}{\rotatebox[origin=c]{90}{PEMS-BAY}}
& DCRNN~\cite{LiYS018} &   4.41 & 6.07 & 10.51\% & 5.95 & 8.45 & 15.36\% & 7.50 & 10.53 & 20.39\% & 5.95 & 8.35 & 15.42\%  \\ 
& \textbf{+ Calibration} & \textbf{4.25} & \textbf{5.93} & \textbf{10.07\%} & \textbf{5.65} & \textbf{8.15} & \textbf{14.37\%} & \textbf{6.94} & \textbf{10.00} & \textbf{18.60\%} & \textbf{5.61} & \textbf{8.03} & \textbf{14.34\%}  \\ 
\cmidrule{3-14}
& STGCN~\cite{YuYZ18} &  6.42 & 7.95 & 15.88\% & 7.50 & 9.44 & 19.39\% & 8.74 & 11.03 & 23.50\% & 7.56 & 9.47 & 19.59\% \\ 
& \textbf{+ Calibration} & \textbf{3.88} & \textbf{5.91} & \textbf{9.55\%} & \textbf{5.41} & \textbf{8.00} & \textbf{14.36\%} & \textbf{6.84} & \textbf{9.78} & \textbf{18.94\%} & \textbf{5.38} & \textbf{7.90} & \textbf{14.29\%} \\ 
\cmidrule{3-14}
& Graph WaveNet~\cite{WuPLJZ19} &  4.36 & 5.89 & 10.45\% & 5.72 & 7.97 & 14.46\% & 6.89 & 9.50 & 18.07\% & 5.66 & 7.79 & 14.33\%  \\ 
& \textbf{+ Calibration} & \textbf{4.28} & \textbf{5.87} & \textbf{10.07\%} & \textbf{5.63} & \textbf{7.93} & \textbf{14.21\%} & \textbf{6.81} & \textbf{9.46} & \textbf{18.02\%} & \textbf{5.57} & \textbf{7.75} & \textbf{14.10\%} \\ 
\cmidrule{3-14}
& STAWnet~\cite{itr2.12044} &  4.38 & 5.96 & 10.34\% & 5.71 & 7.91 & 14.51\% & 6.75 & 9.29 & 18.07\% & 5.61 & 7.72 & 14.31\%  \\ 
& \textbf{+ Calibration} &  \textbf{4.27} & \textbf{5.87} & \textbf{10.07\%} & \textbf{5.62} & \textbf{7.88} & \textbf{14.23\%} & \textbf{6.70} & \textbf{9.27} & \textbf{17.77\%} & \textbf{5.53} & \textbf{7.67} & \textbf{14.02\%} \\ 

\bottomrule
\end{tabular}
\end{center}
% \vspace{-0.1in}
\end{table*}

%% file: 06_Experimental_Results.tex
\section{Experimental Results}
\label{sec_exp_results}

\subsection{Synthetic Dataset}
To validate the correctness of our \method, we first construct a simple synthetic dataset where similar events are occurring at random time steps. This reflects the nature of traffic data which has a similar propagation of congestion in cases of accidents. 

Concretely, the synthetic dataset with the lengths of $10k$ time steps contains a periodic sine wave with a randomly generated zero signal, as depicted in Fig.~\ref{fig_qual_main_results}~(a). 
The period of the sine wave is set to 50 steps and each period with zero values is randomly substituted with a probability of $0.1$ to reflect the traffic dynamics. Similar to the traffic datasets, the synthetic dataset is divided into three parts: the training set, validation set, and test set with a ratio of 7:1:2.
With our synthetic dataset, we examine the correctness of the following two assumptions essential for residual correction: (i) a deep-learning-based prediction model likely generates similar errors in similar types of events, and (ii) when the residuals are correlated, it is possible to improve the performance of the base prediction model by estimating the residuals that will occur, \ie, \emph{how the model fails}.

For the base prediction model, we build a simple sequence-to-sequence model (Seq2seq). Seq2seq is designed to get an input sequence of length 24 and generate predictions on the next 24 steps. The encoder and decoder of Seq2seq are conducted with GRU units with a hidden feature size of 128 and a single recurrent layer.
The output of the decoder is passed to a multi-layer perceptron (MLP) consisting of ReLU activations and fully connected layers of size 128-16-1. The model is trained using an Adam optimizer~\cite{KingmaB14} with a learning rate of 0.001, $\beta_{1} =0.9$, $\beta_{2} =0.999$, and a batch size of 100. The model is trained for 50 epochs and the Z-score normalization is applied to preprocess the input. 

Fig.~\ref{fig_qual_main_results}~(a) shows the predictions of the Seq2seq model, the estimated residuals from our \method, and the calibrated prediction, respectively. As we expected in the first assumption, the prediction model always made similar errors for similar types of events. This implies that the residuals do not occur randomly and are also predictable.
To assess the second assumption, we trained our \method to predict the residuals occurring in Seq2seq. For the synthetic dataset without the graph structure, the spatial-temporal layers in the encoder were replaced with 1D convolutional layers. We set $d_c=32$, $n_c=16$, and $d_e=16$ for the quantization branch in the decoder.
For training, an Adam optimizer with a learning rate of 0.001 and a batch size of 128 was used. Since the length of Seq2seq is $T=24$ steps, our \method gets $24 \times 24$ residuals with the original time series data for the input and outputs the $24 \times 1$ size of the residuals.
Table~\ref{table_synthetic} shows the results of Seq2seq with and without our \method. Our \method is shown to greatly enhance the performance of Seq2seq in every step of the predictions. This indicates that our \method accurately predicts the residuals to occur. Notably, our \method further improves the MAE of Seq2seq by  0.014, 0.036, 0.056, and 0.052 for 1 step, 6 steps, 12 steps, and 24 steps, respectively.
Through a simple simulation, we validate that our assumptions and the proposed method are quite presumable in the time series data. The calibrated results in Fig.~\ref{fig_qual_main_results}~(a) demonstrate that repeated errors can be estimated by our \method. Moreover, the experiments on the synthetic data show that \method allows the model to rapidly adapt to unexpected changes.

\input{Figures/Figure_Qual_Cluster}

\subsection{Traffic Dataset}
\label{sec_exp_results_traffic_dataset}

In this section, we demonstrate that our \method can correct the failure of the existing traffic forecasting models on the METR-LA and PEMS-BAY datasets. 
Our \method is trained with the settings as described in Section~\ref{sec_exp_settings}.

\noindent
\textbf{Residual Correction in Event Situations.}
To reflect the nature of the real-world setting, we examine our \method in regions where critical errors occur. 
Table~\ref{table_high_error} shows the calibration results on time steps where the absolute error of the base forecasting model falls within the top $20\%$. 
Surprisingly, we can observe large gaps in performance before and after calibration. In the event situations on METR-LA, our \method improves the performance of DCRNN by 0.79, 0.62 and 3.32\% for MAE, RMSE, and MAPE, respectively. Even for STAWnet, the most advanced traffic forecasting model, our \method shows an improvement of 0.29, 0.23 and 0.95\% for MAE, RMSE, and MAPE on METR-LA, respectively. Consequently, our \method can calibrate the prediction models more effectively in the cases of critical errors, which highlights the practicality of our \method in real-time traffic forecasting.

As shown in Fig.~\ref{fig_correlation_all}, our ResCAL can accurately estimate the residuals and drastically reduce the autocorrelation of residuals both temporally and spatially. 
Fig.~\ref{fig_qual_main_results}~(b) shows the qualitative results of prediction models with and without our proposed \method on METR-LA.
When the speed drops rapidly (\ie, an anomalous event occurs), STGCN and DCRNN cannot adapt to the changes; thus, they show poor prediction performance in the changed regions (orange lines). For the same case, our \method successfully captures what those models tend to overestimate and corrects them accurately (green lines). 
Consequently, as a model-agnostic add-on module, our proposed \method allows the prediction model to adapt to the fluctuation of data and consistently improves prediction performance regardless of the prediction model.

\input{Tables/Table_Main_Results_one_column}

\noindent
\textbf{Residual Correction in Overall Situations.}
While our residual correction clearly shows its effectiveness in situations where residuals are correlated, we can also examine its performance when the existing models already correctly predict speeds. Here, we examine our method in overall situations for both METR-LA and PEMS-BAY to demonstrate its consistency.
Table~\ref{table_main_results_one_column} shows that our \method consistently improves the baselines on METR-LA and PEMS-BAY.
This indicates that deep-learning-based models generate the correlated residuals even with their large number of parameters, and performance improvement in a high error region also leads to improvement in overall situations. 
For STGCN, our \method achieves average improvements of 0.28 in MAE, 0.41 in RMSE, and 1.01\% in MAPE on METR-LA.
Even with the most recent model STAWnet, our \method shows improvements for MAE, RMSE, and MAPE in both METR-LA and PEMS-BAY.

\noindent
\textbf{Running Time Analysis.}
The average computation time required for calibrating the outputs of the base models (\eg, Graph-WaveNet) was measured in real-time inference. It was tested on METR-LA and 12 sequences of Graph-WaveNet outputs were calibrated.
A PC with an Intel Xeon Silver 4210R 2.40GHz CPU and a Titan RTX GPU was used in our analysis. On average, the calibration was performed within 9.05ms. That is, the running time of ResCAL is fast enough for practical usage. Note that the inference time of our ResCAL does not depend on the base model.  

\subsection{Pattern Analysis and Interpretability}
Here, we show the qualitative results for the residuals assigned in the same categorical variables to examine the role of the quantization branch of \method.
Concretely, we extract the quantized vectors with the dimensions of the number of categorical variables $D=32$ from the Gumbel-Softmax operation and assign the two residuals in the same pattern if both vectors have identical values for 32 categorical variables.
Fig.~\ref{fig_qual_cluster} shows the prediction results of STGCN on METR-LA for the two different input sequences, each from the train dataset and test dataset. The quantized vectors for each pair in Fig.~\ref{fig_qual_cluster}~(a), (b), and (c) belong to the same pattern.
Notable implication comes from the fact that the residuals assigned to the same category show a similar pattern of the sequences. The results for the train dataset (top) and test dataset (bottom) share a similar ground truth with a similar strategy of calibration. That is, our \method can provide the \emph{interpretable evidences} for calibration. This is crucial for real-time traffic forecasting since the calibration of residuals is mostly needed in the case of unintended failures, and case studies are essential for maintenance of the system.

\input{Tables/Table_Tuning_Analysis_one_column}

\subsection{Tuning Analysis}
\label{sec:tuning_analysis}

In the tuning analysis, we show that the performance gain of \method is mostly based on the consideration of residuals rather than on external factors (\eg, introducing additional parameters). As an add-on module, our \method introduces additional parameters compared to solely using a baseline. In addition, \method uses the residuals from previous predictions, thus observing a wide range of input sequences.
For scrutiny, we conducted an experiment to check whether the performance gains can be obtained by simply increasing the model complexity or increasing the input length without using \method. 
We introduced two variations of the baselines: one with larger parameters (LH) and one with longer input sequences (LI). Concretely, we extend the channels of the convolutional layers to increase the number of parameters as much as our \method ($\sim170k$) for LH models and provide $2\times$ longer input sequences for the LI models. Table~\ref{table_tuning_analysis} shows the results of the two variations and the original with our \method on METR-LA. Interestingly, even with the larger parameters and wide ranges of usable data, the baselines cannot reproduce the results of our \method. In Graph WaveNet-LH, introducing the larger parameters seems to improve the performance in terms of MAPE. However, the performance of MAE and RMSE significantly decreased, even compared to the original Graph WaveNet. In contrast, the model with our \method shows a stable performance improvement in all cases.

\input{Figures/Figure_Residual_Cluster}

\subsection{Sanity Check on Clustering}
To validate the reliability of our quantization branch, we performed a sanity check by assessing the patterns with lower occurrences. Fig.~\ref{fig_residual_cluster_two_columns} highlights $5\%$ of the time steps assigned to the least frequent patterns. From this, we can observe that abnormal residuals mainly occur in the selected regions. This demonstrates that our quantization branch can capture minor patterns faithfully without being biased toward dominant events.

%% file: Figures/Figure_Qual_Cluster.tex
\begin{figure*}[t!]
\begin{center}
    \includegraphics[width=1.0\linewidth]{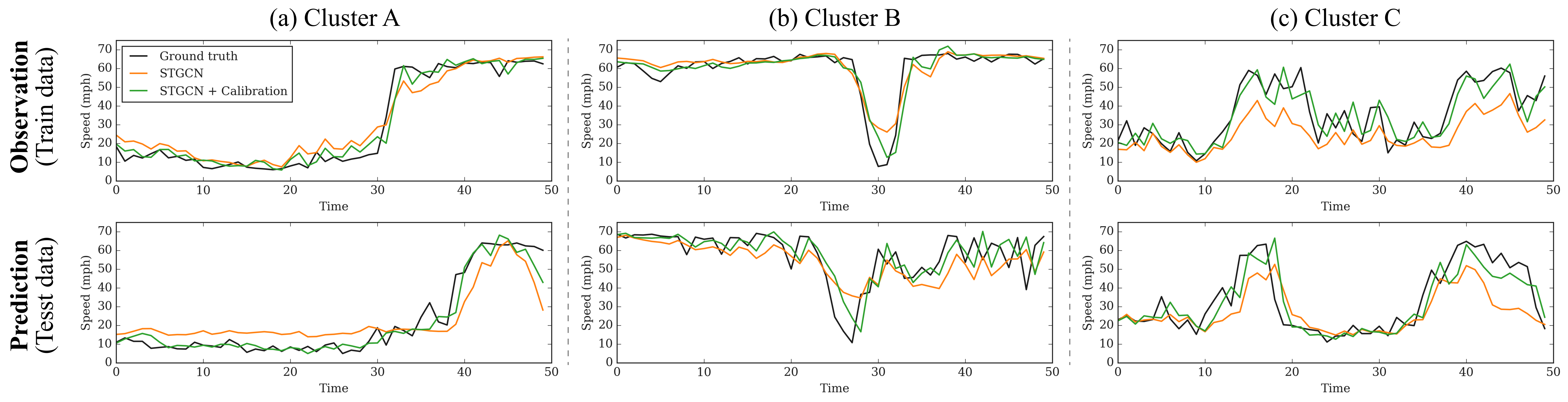}
\end{center}
    \vspace{-0.1in}
    \caption{The calibration results on the train dataset and test dataset of the METR-LA dataset. Each pair in (a), (b), and (c) contains two different results assigned in the same pattern. Events of the same pattern in the train dataset can be interpreted as \emph{evidences} of the calibration of our \method in real-time traffic forecasting.}
    % \vspace{-0.1in}
\label{fig_qual_cluster}
\end{figure*}

%% file: Tables/Table_Main_Results_one_column.tex
\begin{table}[t!]
\begin{center}
\caption{Experimental results on METR-LA and PEMS-BAY. A model with a star denotes that its results were obtained from the original work. Otherwise, the results are reproduced as detailed in Section~\ref{sec_exp_settings}.}
% \vspace{-0.1in}
\label{table_main_results_one_column}
\resizebox{\linewidth}{!}{
\setlength{\tabcolsep}{3pt}
\begin{tabular}[\linewidth]{l l ccc | ccc | ccc}
\toprule
\multirow{2}{*}{Data} & \multirow{2}{*}{Models}  & \multicolumn{3}{c}{15min}   & \multicolumn{3}{c}{30min} & \multicolumn{3}{c}{60min} \\
\cmidrule{3-11}
&&  MAE & RMSE  & MAPE  &  MAE & RMSE  & MAPE  &  MAE & RMSE  & MAPE \\

\midrule\midrule

\multirow{12}{*}{\rotatebox[origin=c]{90}{METR-LA}}   &
ARIMA$^\ast$ &   3.99   &   8.21   &   9.60\%   &   5.15   &  10.45  &   12.70\%   &   6.90   &   13.23   &   17.40\% \\
&FC-LSTM$^\ast$ &   3.44   &   6.30   &   9.60\%   &   3.77   & 7.23  &   10.90\%   &   4.37   &   8.69   &   13.20\%   \\ 
\cmidrule{3-11}
&DCRNN~\cite{LiYS018} &   3.17   &   5.53   &   8.28\%   &   3.53   &   6.33   &   9.63\%   &   4.02   &   7.32   &   11.40\%   \\ 
&\textbf{+ Calibration} &   \textbf{3.05}   &   \textbf{5.32}   &   \textbf{7.81\%}   &   \textbf{3.44}   &   \textbf{6.20}   &   \textbf{9.25\%}   &   \textbf{3.98}   &   \textbf{7.21}   &   \textbf{11.23\%}   \\ 
\cmidrule{3-11}
&STGCN~\cite{YuYZ18} &   3.33   &   5.77   &   8.92\%   &   3.74   &   6.65   &  10.39\%   &   4.32   &   7.72   &   12.52\%   \\ 
&\textbf{+ Calibration} &   \textbf{2.98}   &   \textbf{5.28}   &   \textbf{7.73\%}   &   \textbf{3.47}   &   \textbf{6.26}   &   \textbf{9.46\%}   &   \textbf{4.10}   &   \textbf{7.36}   &   \textbf{11.62\%}   \\ 
\cmidrule{3-11}
&GWNet~\cite{WuPLJZ19} &   \textbf{2.88}   &   5.10   &   {\textbf{7.30\%}}   &   \textbf{3.33}   &   6.03   &   8.93\%   &   3.91   &   7.01   &  10.84\%   \\ 
&\textbf{+ Calibration} &   \textbf{2.88}   &   \textbf{5.09}   &   {\textbf{7.30\%}}   &   \textbf{3.33}   &   \textbf{6.02}   &   \textbf{8.85\%}   &   \textbf{3.87}   &   \textbf{6.99}   &   \textbf{10.66\%}   \\ 
\cmidrule{3-11}
&STAWnet~\cite{itr2.12044} &   2.87   &   5.15   &   7.44\%   &   3.28   &   6.03   &   8.94\%   &   3.78   &  6.97  &   {\textbf{10.55\%}}   \\ 
&\textbf{+ Calibration} &  {\textbf{2.85}}   &   {\textbf{5.08}}   &  {\textbf{7.30\%}}   &   {\textbf{3.27}}   &   {\textbf{5.98}}   &   {\textbf{8.81\%}}   &   {\textbf{3.77}}   &   {\textbf{6.94}}   &   10.56\%   \\ 

\midrule\midrule

\multirow{12}{*}{\rotatebox[origin=c]{90}{PEMS-BAY}}
&ARIMA$^\ast$ &  1.62   &   3.30   &   3.50\%   &   2.33   &   4.76   &   5.40\%   &   3.38   &   6.50   &   8.30\% \\
&FC-LSTM$^\ast$ &   2.05   &   4.19   &   4.80\%   &   2.20   &   4.55   &   5.20\%   &   2.37   &   4.96   &   5.70\%   \\ 
\cmidrule{3-11}
&DCRNN~\cite{LiYS018} &   1.40   &   2.80   &   2.95\%   &   1.79   &   3.87   &   4.04\%   &   2.21   &   4.81   &   5.22\%   \\ 
&\textbf{+ Calibration} &   \textbf{1.37}   &   \textbf{2.75}   &   \textbf{2.87\%}   &   \textbf{1.76}   &   \textbf{3.77}   &   \textbf{3.91\%}   &   \textbf{2.16}   &   \textbf{4.64}   &   \textbf{5.00\%}   \\ 
\cmidrule{3-11}
&STGCN~\cite{YuYZ18} &   2.15   &   3.75   &   4.57\%   &   2.42   &   4.40   &   5.35\%   &   2.76   &   5.12   &   6.31\%   \\ 
&\textbf{+ Calibration} &   \textbf{1.46}   &   \textbf{2.87}   &   \textbf{3.07\%}   &   \textbf{1.85}   &   \textbf{3.79}   &   \textbf{4.16\%}   &   \textbf{2.27}   &   \textbf{4.61}   &   \textbf{5.29\%}   \\ 
\cmidrule{3-11}
&GWNet~\cite{WuPLJZ19} &   1.37   &   {\textbf{2.72}}   &   2.90\%   &   1.72   &   3.65   &   3.82\%   &  \textbf{2.03}   &   4.35   &   \textbf{4.67\%}   \\ 
&\textbf{+ Calibration} &   {\textbf{1.36}}   &   {\textbf{2.72}}   &   {\textbf{2.84\%}}   &   {\textbf{1.71}}   &   \textbf{3.64}   &   {\textbf{3.79\%}}   &   \textbf{2.03}   &   \textbf{4.33}   &   4.68\%   \\ 
\cmidrule{3-11}
&STAWnet~\cite{itr2.12044} &   1.37   &   2.75   &   2.88\%   &   1.72   &   3.63   &   3.83\%   &   {\textbf{2.01}}   &   {\textbf{4.25}}   &   4.68\%   \\ 
&\textbf{+ Calibration} &   {\textbf{1.36}}   &   {\textbf{2.72}}   &   \textbf{2.86\%}   &  {\textbf{1.71}}   &   {\textbf{3.62}}   &   \textbf{3.81\%}   &   {\textbf{2.01}}   &   {\textbf{4.25}}   &   {\textbf{4.64\%}}   \\ 

\bottomrule
\end{tabular}
}
\vspace{-0.1in}
\end{center}
\end{table}

%% file: Tables/Table_Tuning_Analysis_one_column.tex
\begin{table}[t!]
\begin{center}
\caption{Tuning analysis of \method with Graph WaveNet.}
\vspace{-0.1in}
\label{table_tuning_analysis}
\resizebox{\linewidth}{!}{
\setlength{\tabcolsep}{3pt}
\begin{tabular}[\linewidth]{l ccc | ccc | ccc}
\toprule
\multirow{2}{*}{Models}  & \multicolumn{3}{c}{15min}   & \multicolumn{3}{c}{30min} & \multicolumn{3}{c}{60min}  \\
\cmidrule{2-10}
&  MAE & RMSE  & MAPE  &  MAE & RMSE  & MAPE  &  MAE & RMSE  & MAPE \\

\midrule\midrule

STGCN~\cite{YuYZ18} &   \underline{3.33}   &   \underline{5.77}   &   \underline{8.92\% }  &   \underline{3.74 }  &  \underline{ 6.65 }  &  \underline{10.39\%}   &  \underline{ 4.32 }  &   \underline{7.72}   &  \underline{ 12.52\%} \\
\midrule
STGCN$-\textrm{LH}$ & 3.43   &  5.87 &  9.09\% & 3.85  & 6.85 &  10.84\%  & 4.40 & 7.95 &  12.94\% \\
STGCN$-\textrm{LI}$ &  3.46  &  5.84 &   9.00\%  &  3.90   & 6.75 & 10.56\%  &  4.46  &  7.87 &  12.77\% \\
STGCN + Ours  & \textbf{2.98}   &   \textbf{5.28}   &   \textbf{7.73\%}   &   \textbf{3.47}   &   \textbf{6.26}   &   \textbf{9.46\%}   &   \textbf{4.10}   &   \textbf{7.36}   &   \textbf{11.62\%}  \\
\midrule\midrule

Graph-WaveNet~\cite{WuPLJZ19} &  \textbf{2.88}   &   \underline{5.10}   &   \underline{7.30\%}   &   \textbf{3.33}   &  \underline{6.03}   &   8.93\%   &   \underline{3.91}   &   \underline{7.01}   &  10.84\% \\
\midrule
Graph WaveNet$-\textrm{LH}$ &  \underline{2.90}   &   5.13   &   \textbf{7.23\%}   &   \underline{3.37}   &   6.11   &   \textbf{8.72\%}  &   4.01   &   7.22   &   \textbf{10.57\%} \\
Graph WaveNet$-\textrm{LI}$ &   3.31  &   5.55   &   8.43\%   &   3.62   &   6.29   &   9.66\%   &   4.13   &   7.17   &   11.26\% \\
Graph WaveNet + Ours  &   \textbf{2.88}   &   \textbf{5.09}   &   \underline{7.30\%}   &   \textbf{3.33}   &   \textbf{6.02}   &   \underline{8.85\%}   &   \textbf{3.87}   &   \textbf{6.99}   &  \underline{10.66\%} \\

\bottomrule
\end{tabular}
}
\end{center}
\vspace{-0.2in}
\end{table}

%% file: Figures/Figure_Residual_Cluster.tex
\begin{figure}[t!]
\begin{center}
    \includegraphics[width=1.0\linewidth]{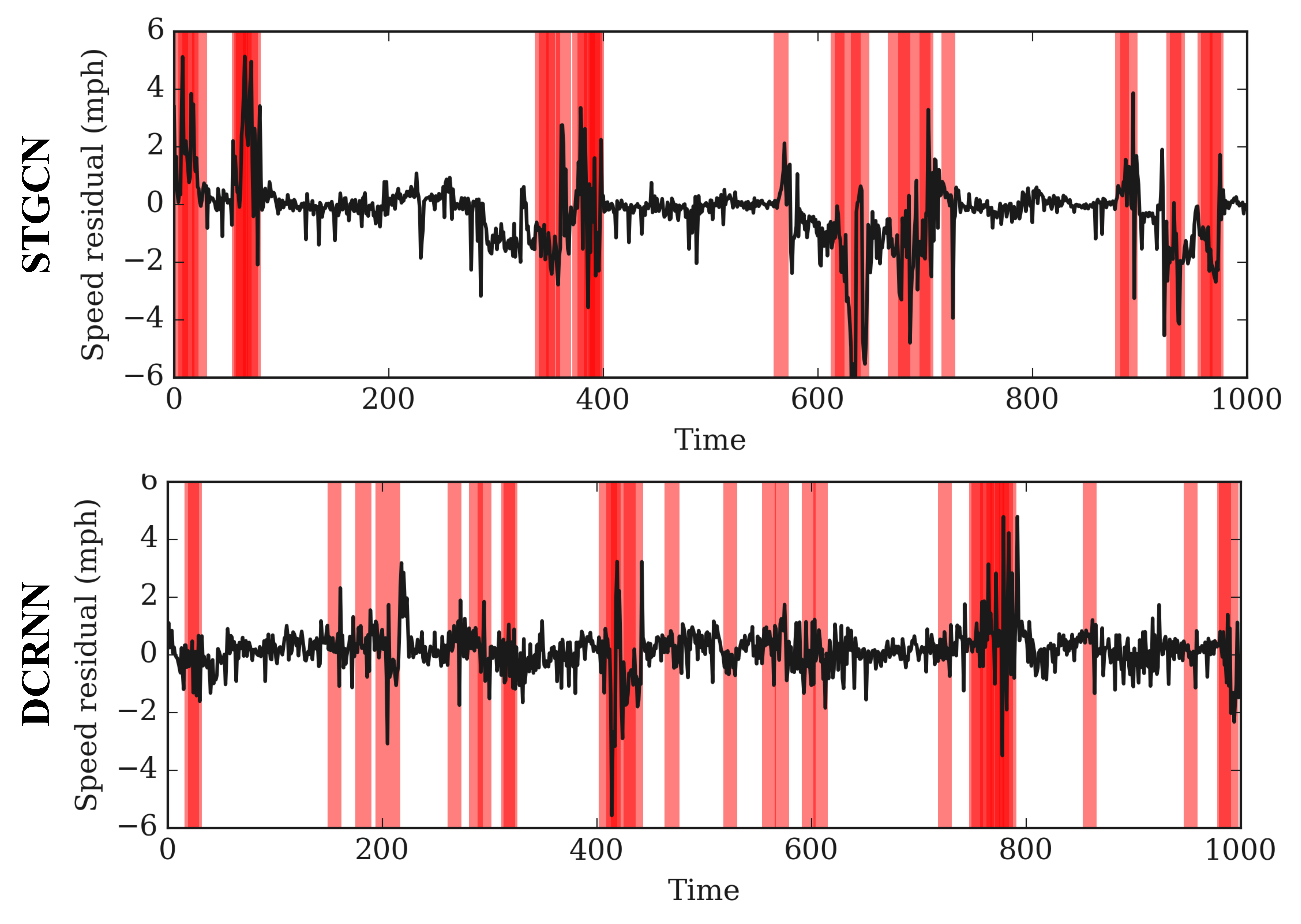}
\end{center}
    \vspace{-0.2in}
    \caption{The residual patterns of the baseline models on METR-LA. The regions of 5\% minor patterns are highlighted in red.}
\label{fig_residual_cluster_two_columns}
\vspace{-0.1in}
\end{figure}

%% file: 07_Conclusion.tex
\section{Conclusion}

In this work, we introduce a real-time setting in traffic forecasting and show that the residuals occurring from deep-learning-based models are highly correlated both spatially and temporally. To fully consider the autocorrelation of the residuals, we present a model-agnostic add-on module named \method, which calibrates the predictions by estimating the residuals. 
Without having to fix the original architectures, which may entail high computational cost, our \method module can effectively capture spatial-temporal dependencies of the residuals. 
On METR-LA and PEMS-BAY, our \method consistently shows performance improvements for the existing forecasting models in event situations. 
Furthermore, we demonstrate the high practicality of our \method in real-time traffic forecasting by providing the interpretability for calibration and effectively calibrating the predictions with significant errors. While we focus on traffic datasets, our proposed approach can be freely adopted to various tasks with autocorrleated residuals. 
We hope that our findings will shed light on future research for other forecasting problems as well as traffic forecasting.

%% file: main.bbl
%%% -*-BibTeX-*-
%%% Do NOT edit. File created by BibTeX with style
%%% ACM-Reference-Format-Journals [18-Jan-2012].

\begin{thebibliography}{30}

%%% ====================================================================
%%% NOTE TO THE USER: you can override these defaults by providing
%%% customized versions of any of these macros before the \bibliography
%%% command.  Each of them MUST provide its own final punctuation,
%%% except for \shownote{}, \showDOI{}, and \showURL{}.  The latter two
%%% do not use final punctuation, in order to avoid confusing it with
%%% the Web address.
%%%
%%% To suppress output of a particular field, define its macro to expand
%%% to an empty string, or better, \unskip, like this:
%%%
%%% \newcommand{\showDOI}[1]{\unskip}   % LaTeX syntax
%%%
%%% \def \showDOI #1{\unskip}           % plain TeX syntax
%%%
%%% ====================================================================

\ifx \showCODEN    \undefined \def \showCODEN     #1{\unskip}     \fi
\ifx \showDOI      \undefined \def \showDOI       #1{#1}\fi
\ifx \showISBNx    \undefined \def \showISBNx     #1{\unskip}     \fi
\ifx \showISBNxiii \undefined \def \showISBNxiii  #1{\unskip}     \fi
\ifx \showISSN     \undefined \def \showISSN      #1{\unskip}     \fi
\ifx \showLCCN     \undefined \def \showLCCN      #1{\unskip}     \fi
\ifx \shownote     \undefined \def \shownote      #1{#1}          \fi
\ifx \showarticletitle \undefined \def \showarticletitle #1{#1}   \fi
\ifx \showURL      \undefined \def \showURL       {\relax}        \fi
% The following commands are used for tagged output and should be
% invisible to TeX
\providecommand\bibfield[2]{#2}
\providecommand\bibinfo[2]{#2}
\providecommand\natexlab[1]{#1}
\providecommand\showeprint[2][]{arXiv:#2}

\bibitem[Atwood and Towsley(2016)]%
        {AtwoodT16}
\bibfield{author}{\bibinfo{person}{James Atwood} {and} \bibinfo{person}{Don
  Towsley}.} \bibinfo{year}{2016}\natexlab{}.
\newblock \showarticletitle{Diffusion-Convolutional Neural Networks}. In
  \bibinfo{booktitle}{\emph{Proc. the Advances in Neural Information Processing
  Systems (NeurIPS)}}.
\newblock


\bibitem[Box et~al\mbox{.}(1976)]%
        {box1976time}
\bibfield{author}{\bibinfo{person}{George~EP Box}, \bibinfo{person}{Gwilym~M
  Jenkins}, {and} \bibinfo{person}{Gregory~C Reinsel}.}
  \bibinfo{year}{1976}\natexlab{}.
\newblock \bibinfo{title}{Time series analysis prediction and control}.
\newblock
\newblock


\bibitem[Chen et~al\mbox{.}(2015)]%
        {chen2015xgboost}
\bibfield{author}{\bibinfo{person}{Tianqi Chen}, \bibinfo{person}{Tong He},
  \bibinfo{person}{Michael Benesty}, \bibinfo{person}{Vadim Khotilovich},
  \bibinfo{person}{Yuan Tang}, \bibinfo{person}{Hyunsu Cho}, {et~al\mbox{.}}}
  \bibinfo{year}{2015}\natexlab{}.
\newblock \showarticletitle{Xgboost: extreme gradient boosting}.
\newblock \bibinfo{journal}{\emph{R package version 0.4-2}}
  \bibinfo{volume}{1}, \bibinfo{number}{4} (\bibinfo{year}{2015}),
  \bibinfo{pages}{1--4}.
\newblock


\bibitem[Chen et~al\mbox{.}(2016)]%
        {ChenCDHSSA16}
\bibfield{author}{\bibinfo{person}{Xi Chen}, \bibinfo{person}{Yan Duan},
  \bibinfo{person}{Rein Houthooft}, \bibinfo{person}{John Schulman},
  \bibinfo{person}{Ilya Sutskever}, {and} \bibinfo{person}{Pieter Abbeel}.}
  \bibinfo{year}{2016}\natexlab{}.
\newblock \showarticletitle{InfoGAN: Interpretable Representation Learning by
  Information Maximizing Generative Adversarial Nets}. In
  \bibinfo{booktitle}{\emph{Proc. the Advances in Neural Information Processing
  Systems (NeurIPS)}}.
\newblock


\bibitem[Cheng et~al\mbox{.}(2018)]%
        {ChengZZX18}
\bibfield{author}{\bibinfo{person}{Xingyi Cheng}, \bibinfo{person}{Ruiqing
  Zhang}, \bibinfo{person}{Jie Zhou}, {and} \bibinfo{person}{Wei Xu}.}
  \bibinfo{year}{2018}\natexlab{}.
\newblock \showarticletitle{DeepTransport: Learning Spatial-Temporal Dependency
  for Traffic Condition Forecasting}. In \bibinfo{booktitle}{\emph{2018
  International Joint Conference on Neural Networks, {IJCNN} 2018, Rio de
  Janeiro, Brazil, July 8-13, 2018}}.
\newblock


\bibitem[Friedman(2001)]%
        {friedman2001greedy}
\bibfield{author}{\bibinfo{person}{Jerome~H Friedman}.}
  \bibinfo{year}{2001}\natexlab{}.
\newblock \showarticletitle{Greedy function approximation: a gradient boosting
  machine}.
\newblock \bibinfo{journal}{\emph{Annals of statistics}}
  (\bibinfo{year}{2001}), \bibinfo{pages}{1189--1232}.
\newblock


\bibitem[Gray(1984)]%
        {gray1984vector}
\bibfield{author}{\bibinfo{person}{Robert Gray}.}
  \bibinfo{year}{1984}\natexlab{}.
\newblock \showarticletitle{Vector quantization}.
\newblock \bibinfo{journal}{\emph{IEEE Assp Magazine}} \bibinfo{volume}{1},
  \bibinfo{number}{2} (\bibinfo{year}{1984}), \bibinfo{pages}{4--29}.
\newblock


\bibitem[Hochreiter and Schmidhuber(1997)]%
        {HochreiterS97}
\bibfield{author}{\bibinfo{person}{Sepp Hochreiter} {and}
  \bibinfo{person}{J{\"{u}}rgen Schmidhuber}.} \bibinfo{year}{1997}\natexlab{}.
\newblock \showarticletitle{Long Short-Term Memory}.
\newblock \bibinfo{journal}{\emph{Neural Comput.}} (\bibinfo{year}{1997}).
\newblock


\bibitem[Huang et~al\mbox{.}(2021)]%
        {HuangHSLB21}
\bibfield{author}{\bibinfo{person}{Qian Huang}, \bibinfo{person}{Horace He},
  \bibinfo{person}{Abhay Singh}, \bibinfo{person}{Ser{-}Nam Lim}, {and}
  \bibinfo{person}{Austin~R. Benson}.} \bibinfo{year}{2021}\natexlab{}.
\newblock \showarticletitle{Combining Label Propagation and Simple Models
  out-performs Graph Neural Networks}. In \bibinfo{booktitle}{\emph{Proc. the
  International Conference on Learning Representations (ICLR)}}.
\newblock


\bibitem[Hyndman and Athanasopoulos(2018)]%
        {hyndman2018forecasting}
\bibfield{author}{\bibinfo{person}{Rob~J Hyndman} {and} \bibinfo{person}{George
  Athanasopoulos}.} \bibinfo{year}{2018}\natexlab{}.
\newblock \bibinfo{booktitle}{\emph{Forecasting: principles and practice}}.
\newblock \bibinfo{publisher}{OTexts}.
\newblock


\bibitem[Jagadish et~al\mbox{.}(2014)]%
        {JagadishGLPPRS14}
\bibfield{author}{\bibinfo{person}{H.~V. Jagadish}, \bibinfo{person}{Johannes
  Gehrke}, \bibinfo{person}{Alexandros Labrinidis}, \bibinfo{person}{Yannis
  Papakonstantinou}, \bibinfo{person}{Jignesh~M. Patel}, \bibinfo{person}{Raghu
  Ramakrishnan}, {and} \bibinfo{person}{Cyrus Shahabi}.}
  \bibinfo{year}{2014}\natexlab{}.
\newblock \showarticletitle{Big data and its technical challenges}.
\newblock \bibinfo{journal}{\emph{Commun. {ACM}}} (\bibinfo{year}{2014}).
\newblock


\bibitem[Jang et~al\mbox{.}(2017)]%
        {JangGP17}
\bibfield{author}{\bibinfo{person}{Eric Jang}, \bibinfo{person}{Shixiang Gu},
  {and} \bibinfo{person}{Ben Poole}.} \bibinfo{year}{2017}\natexlab{}.
\newblock \showarticletitle{Categorical Reparameterization with
  Gumbel-Softmax}. In \bibinfo{booktitle}{\emph{Proc. the International
  Conference on Learning Representations (ICLR)}}.
\newblock


\bibitem[Jia and Benson(2020)]%
        {JiaB20}
\bibfield{author}{\bibinfo{person}{Junteng Jia} {and}
  \bibinfo{person}{Austin~R. Benson}.} \bibinfo{year}{2020}\natexlab{}.
\newblock \showarticletitle{Residual Correlation in Graph Neural Network
  Regression}. In \bibinfo{booktitle}{\emph{Proc. the ACM SIGKDD International
  Conference on Knowledge Discovery and Data Mining (KDD)}}.
\newblock


\bibitem[Ke et~al\mbox{.}(2017)]%
        {KeMFWCMYL17}
\bibfield{author}{\bibinfo{person}{Guolin Ke}, \bibinfo{person}{Qi Meng},
  \bibinfo{person}{Thomas Finley}, \bibinfo{person}{Taifeng Wang},
  \bibinfo{person}{Wei Chen}, \bibinfo{person}{Weidong Ma},
  \bibinfo{person}{Qiwei Ye}, {and} \bibinfo{person}{Tie{-}Yan Liu}.}
  \bibinfo{year}{2017}\natexlab{}.
\newblock \showarticletitle{LightGBM: {A} Highly Efficient Gradient Boosting
  Decision Tree}. In \bibinfo{booktitle}{\emph{Proc. the Advances in Neural
  Information Processing Systems (NeurIPS)}}.
\newblock


\bibitem[Kingma and Ba(2015)]%
        {KingmaB14}
\bibfield{author}{\bibinfo{person}{Diederik~P. Kingma} {and}
  \bibinfo{person}{Jimmy Ba}.} \bibinfo{year}{2015}\natexlab{}.
\newblock \showarticletitle{Adam: {A} Method for Stochastic Optimization}. In
  \bibinfo{booktitle}{\emph{Proc. the International Conference on Learning
  Representations (ICLR)}}.
\newblock


\bibitem[Kipf and Welling(2017)]%
        {KipfW17}
\bibfield{author}{\bibinfo{person}{Thomas~N. Kipf} {and} \bibinfo{person}{Max
  Welling}.} \bibinfo{year}{2017}\natexlab{}.
\newblock \showarticletitle{Semi-Supervised Classification with Graph
  Convolutional Networks}. In \bibinfo{booktitle}{\emph{Proc. the International
  Conference on Learning Representations (ICLR)}}.
\newblock


\bibitem[Li et~al\mbox{.}(2018)]%
        {LiYS018}
\bibfield{author}{\bibinfo{person}{Yaguang Li}, \bibinfo{person}{Rose Yu},
  \bibinfo{person}{Cyrus Shahabi}, {and} \bibinfo{person}{Yan Liu}.}
  \bibinfo{year}{2018}\natexlab{}.
\newblock \showarticletitle{Diffusion Convolutional Recurrent Neural Network:
  Data-Driven Traffic Forecasting}. In \bibinfo{booktitle}{\emph{Proc. the
  International Conference on Learning Representations (ICLR)}}.
\newblock


\bibitem[Park et~al\mbox{.}(2020)]%
        {ParkLBTJKKC20}
\bibfield{author}{\bibinfo{person}{Cheonbok Park}, \bibinfo{person}{Chunggi
  Lee}, \bibinfo{person}{Hyojin Bahng}, \bibinfo{person}{Yunwon Tae},
  \bibinfo{person}{Seungmin Jin}, \bibinfo{person}{Kihwan Kim},
  \bibinfo{person}{Sungahn Ko}, {and} \bibinfo{person}{Jaegul Choo}.}
  \bibinfo{year}{2020}\natexlab{}.
\newblock \showarticletitle{{ST-GRAT:} {A} Novel Spatio-temporal Graph
  Attention Networks for Accurately Forecasting Dynamically Changing Road
  Speed}. In \bibinfo{booktitle}{\emph{Proc. the ACM Conference on Information
  and Knowledge Management (CIKM)}}.
\newblock


\bibitem[Shuman et~al\mbox{.}(2013)]%
        {ShumanNFOV13}
\bibfield{author}{\bibinfo{person}{David~I. Shuman}, \bibinfo{person}{Sunil~K.
  Narang}, \bibinfo{person}{Pascal Frossard}, \bibinfo{person}{Antonio Ortega},
  {and} \bibinfo{person}{Pierre Vandergheynst}.}
  \bibinfo{year}{2013}\natexlab{}.
\newblock \showarticletitle{The Emerging Field of Signal Processing on Graphs:
  Extending High-Dimensional Data Analysis to Networks and Other Irregular
  Domains}.
\newblock \bibinfo{journal}{\emph{{IEEE} Signal Process. Mag.}}
  \bibinfo{volume}{30}, \bibinfo{number}{3} (\bibinfo{year}{2013}),
  \bibinfo{pages}{83--98}.
\newblock


\bibitem[Sutskever et~al\mbox{.}(2014)]%
        {SutskeverVL14}
\bibfield{author}{\bibinfo{person}{Ilya Sutskever}, \bibinfo{person}{Oriol
  Vinyals}, {and} \bibinfo{person}{Quoc~V. Le}.}
  \bibinfo{year}{2014}\natexlab{}.
\newblock \showarticletitle{Sequence to Sequence Learning with Neural
  Networks}. In \bibinfo{booktitle}{\emph{Proc. the Advances in Neural
  Information Processing Systems (NeurIPS)}},
  \bibfield{editor}{\bibinfo{person}{Zoubin Ghahramani}, \bibinfo{person}{Max
  Welling}, \bibinfo{person}{Corinna Cortes}, \bibinfo{person}{Neil~D.
  Lawrence}, {and} \bibinfo{person}{Kilian~Q. Weinberger}} (Eds.).
\newblock


\bibitem[Tian and Chan(2021)]%
        {itr2.12044}
\bibfield{author}{\bibinfo{person}{Chenyu Tian} {and} \bibinfo{person}{Wai
  Kin~(Victor) Chan}.} \bibinfo{year}{2021}\natexlab{}.
\newblock \showarticletitle{Spatial-temporal attention wavenet: A deep learning
  framework for traffic prediction considering spatial-temporal dependencies}.
\newblock \bibinfo{journal}{\emph{IET Intelligent Transport Systems}}
  (\bibinfo{year}{2021}).
\newblock


\bibitem[van~den Oord et~al\mbox{.}(2016)]%
        {OordDZSVGKSK16}
\bibfield{author}{\bibinfo{person}{A{\"{a}}ron van~den Oord},
  \bibinfo{person}{Sander Dieleman}, \bibinfo{person}{Heiga Zen},
  \bibinfo{person}{Karen Simonyan}, \bibinfo{person}{Oriol Vinyals},
  \bibinfo{person}{Alex Graves}, \bibinfo{person}{Nal Kalchbrenner},
  \bibinfo{person}{Andrew~W. Senior}, {and} \bibinfo{person}{Koray
  Kavukcuoglu}.} \bibinfo{year}{2016}\natexlab{}.
\newblock \showarticletitle{WaveNet: {A} Generative Model for Raw Audio}.
\newblock \bibinfo{journal}{\emph{CoRR}}  \bibinfo{volume}{abs/1609.03499}
  (\bibinfo{year}{2016}).
\newblock


\bibitem[van~den Oord et~al\mbox{.}(2017)]%
        {OordVK17}
\bibfield{author}{\bibinfo{person}{A{\"{a}}ron van~den Oord},
  \bibinfo{person}{Oriol Vinyals}, {and} \bibinfo{person}{Koray Kavukcuoglu}.}
  \bibinfo{year}{2017}\natexlab{}.
\newblock \showarticletitle{Neural Discrete Representation Learning}. In
  \bibinfo{booktitle}{\emph{Proc. the Advances in Neural Information Processing
  Systems (NeurIPS)}}.
\newblock


\bibitem[Vaswani et~al\mbox{.}(2017)]%
        {VaswaniSPUJGKP17}
\bibfield{author}{\bibinfo{person}{Ashish Vaswani}, \bibinfo{person}{Noam
  Shazeer}, \bibinfo{person}{Niki Parmar}, \bibinfo{person}{Jakob Uszkoreit},
  \bibinfo{person}{Llion Jones}, \bibinfo{person}{Aidan~N. Gomez},
  \bibinfo{person}{Lukasz Kaiser}, {and} \bibinfo{person}{Illia Polosukhin}.}
  \bibinfo{year}{2017}\natexlab{}.
\newblock \showarticletitle{Attention is All you Need}. In
  \bibinfo{booktitle}{\emph{Proc. the Advances in Neural Information Processing
  Systems (NeurIPS)}}.
\newblock


\bibitem[Wu and Tan(2016)]%
        {WuT16}
\bibfield{author}{\bibinfo{person}{Yuankai Wu} {and} \bibinfo{person}{Huachun
  Tan}.} \bibinfo{year}{2016}\natexlab{}.
\newblock \showarticletitle{Short-term traffic flow forecasting with
  spatial-temporal correlation in a hybrid deep learning framework}.
\newblock \bibinfo{journal}{\emph{CoRR}}  \bibinfo{volume}{abs/1612.01022}
  (\bibinfo{year}{2016}).
\newblock


\bibitem[Wu et~al\mbox{.}(2019)]%
        {WuPLJZ19}
\bibfield{author}{\bibinfo{person}{Zonghan Wu}, \bibinfo{person}{Shirui Pan},
  \bibinfo{person}{Guodong Long}, \bibinfo{person}{Jing Jiang}, {and}
  \bibinfo{person}{Chengqi Zhang}.} \bibinfo{year}{2019}\natexlab{}.
\newblock \showarticletitle{Graph WaveNet for Deep Spatial-Temporal Graph
  Modeling}. In \bibinfo{booktitle}{\emph{Proc. the International Joint
  Conference on Artificial Intelligence (IJCAI)}},
  \bibfield{editor}{\bibinfo{person}{Sarit Kraus}} (Ed.).
\newblock


\bibitem[Yu et~al\mbox{.}(2018)]%
        {YuYZ18}
\bibfield{author}{\bibinfo{person}{Bing Yu}, \bibinfo{person}{Haoteng Yin},
  {and} \bibinfo{person}{Zhanxing Zhu}.} \bibinfo{year}{2018}\natexlab{}.
\newblock \showarticletitle{Spatio-Temporal Graph Convolutional Networks: {A}
  Deep Learning Framework for Traffic Forecasting}. In
  \bibinfo{booktitle}{\emph{Proc. the International Joint Conference on
  Artificial Intelligence (IJCAI)}}.
\newblock


\bibitem[Zhang et~al\mbox{.}(2018)]%
        {ZhangSXMKY18}
\bibfield{author}{\bibinfo{person}{Jiani Zhang}, \bibinfo{person}{Xingjian
  Shi}, \bibinfo{person}{Junyuan Xie}, \bibinfo{person}{Hao Ma},
  \bibinfo{person}{Irwin King}, {and} \bibinfo{person}{Dit{-}Yan Yeung}.}
  \bibinfo{year}{2018}\natexlab{}.
\newblock \showarticletitle{GaAN: Gated Attention Networks for Learning on
  Large and Spatiotemporal Graphs}. In \bibinfo{booktitle}{\emph{Proc. the
  Conference on Uncertainty in Artificial Intelligence (UAI)}}.
\newblock


\bibitem[Zhang et~al\mbox{.}(2016)]%
        {ZhangZQLY16}
\bibfield{author}{\bibinfo{person}{Junbo Zhang}, \bibinfo{person}{Yu Zheng},
  \bibinfo{person}{Dekang Qi}, \bibinfo{person}{Ruiyuan Li}, {and}
  \bibinfo{person}{Xiuwen Yi}.} \bibinfo{year}{2016}\natexlab{}.
\newblock \showarticletitle{DNN-based prediction model for spatio-temporal
  data}. In \bibinfo{booktitle}{\emph{Proceedings of the 24th {ACM}
  {SIGSPATIAL} International Conference on Advances in Geographic Information
  Systems, {GIS} 2016, Burlingame, California, USA, October 31 - November 3,
  2016}}.
\newblock


\bibitem[Zheng et~al\mbox{.}(2020)]%
        {ZhengFW020}
\bibfield{author}{\bibinfo{person}{Chuanpan Zheng}, \bibinfo{person}{Xiaoliang
  Fan}, \bibinfo{person}{Cheng Wang}, {and} \bibinfo{person}{Jianzhong Qi}.}
  \bibinfo{year}{2020}\natexlab{}.
\newblock \showarticletitle{{GMAN:} {A} Graph Multi-Attention Network for
  Traffic Prediction}. \bibinfo{publisher}{Proc. the AAAI Conference on
  Artificial Intelligence (AAAI)}.
\newblock


\end{thebibliography}
